\documentclass[11pt]{article}

\usepackage[utf8]{inputenc} %
\usepackage[T1]{fontenc}    %
\usepackage[english]{babel}
\usepackage[numbers,sort&compress]{natbib}
\usepackage{url}            %
\usepackage{authblk}
\usepackage{booktabs}       %
\usepackage{amsfonts}       %
\usepackage{nicefrac}       %
\usepackage{microtype}      %
\usepackage[table]{xcolor}         %
\usepackage{graphicx}
\usepackage{amsmath,bm}
\usepackage{subcaption}

\usepackage{wrapfig, blindtext}
\usepackage{fancyhdr}
\usepackage{lipsum}
\usepackage{relsize}
\usepackage[export]{adjustbox}
\usepackage{enumitem}       %
\usepackage{todonotes}
\usepackage{amssymb}
\usepackage{caption}
\usepackage{titlesec}
\usepackage{afterpage}
\usepackage{numprint}
\usepackage{soul}
\usepackage[a4paper,
            bindingoffset=0.2in,
            left=1in,
            right=1in,
            top=1in,
            bottom=1in,
            footskip=.25in]{geometry}
\usepackage{changepage}
\usepackage{multirow}

\DeclareCaptionFormat{myformat}{\textbf{#1#2 $|$} #3}
 \renewcommand*{\Affilfont}{\normalsize\normalfont}
\titlespacing*{\section}
{0pt}{5.5ex plus 1ex minus .2ex}{4.3ex plus .2ex}
\titlespacing*{\subsection}
{0pt}{5.5ex plus 1ex minus .2ex}{4.3ex plus .2ex}
\DeclareUnicodeCharacter{00B0}{ }

\newcommand{\themethod}{LPM}
\newcommand{\reploglek}{Replogle et al. (K562)}
\newcommand{\reploglerpe}{Replogle et al. (RPE1)}

\newcommand{\rnn}[1]{{#1}}

\usepackage{color}
\usepackage{tcolorbox}
\usepackage{csquotes}
\tcbset{width=0.9\textwidth,boxrule=0pt,colback=red, arc=0pt, auto outer arc,left=0pt,right=0pt,boxsep=5pt}
\usepackage{hyperref}
\usepackage[capitalise]{cleveref}

\newcommand{\papertitle}{In-silico biological discovery with large perturbation models}
\title{\textbf{\papertitle}}
\begin{document}
\author[1,a,*]{{\DJ}or{\dj}e Miladinovi{\'c}}
\author[1,2,a]{Tobias Höppe}
\author[1]{Mathieu Chevalley}
\author[1]{Andreas Georgiou}
\author[1]{Lachlan Stuart}
\author[1]{Arash Mehrjou}
\author[1]{Marcus Bantscheff}
\author[3]{Bernhard Schölkopf}
\author[1,*]{Patrick Schwab}
\affil[1]{GSK plc, Zug, Switzerland}
\affil[2]{Helmholtz Munich, Germany}
\affil[3]{Max Planck Institute for Intelligent Systems \& ELLIS Institute, Tübingen, Germany}
\affil[a]{Joint first authors}
\affil[*]{Corresponding authors}

\date{}
\setcounter{Maxaffil}{0}
\renewcommand\Affilfont{\itshape\small}
\newcommand{\ManuscriptFormat}{nature} %

\maketitle
\thispagestyle{fancy}
\pagestyle{fancy}
\rhead[RO,LE]{prepublication draft}

\begin{adjustwidth*}{1.5cm}{1.5cm}
\section*{Abstract}

Data generated in perturbation experiments link perturbations to the changes they elicit and therefore contain information relevant to numerous biological discovery tasks -- from understanding the relationships between biological entities to developing therapeutics.
However, these data encompass diverse perturbations and readouts, and the complex dependence of experimental outcomes on their biological context makes it challenging to integrate insights across experiments.
\rnn{Here, we present the Large Perturbation Model (\themethod{}), a deep-learning model that integrates multiple, heterogeneous perturbation experiments by representing perturbation, readout, and context as disentangled dimensions.}
\themethod{} outperforms existing methods across multiple biological discovery tasks, including in predicting post-perturbation transcriptomes of unseen experiments, identifying shared molecular mechanisms of action between chemical and genetic perturbations, and facilitating the inference of gene-gene interaction networks.
\themethod{} learns meaningful joint representations of perturbations, readouts and contexts, enables the study of biological relationships in silico, and could considerably accelerate the derivation of  insights from pooled perturbation experiments.
 \end{adjustwidth*}
\newpage{}

\section{Introduction}
\label{sec:intro}

Perturbation experiments play a central role in elucidating the underlying causal mechanisms that govern the behaviours of biological systems \cite{meinshausen2016methods,rubin2019coupled,tejada2023causal}. Controlled perturbation experiments measure changes in experimental readouts, such as the number of specific transcripts observed, resulting from introducing perturbations to biological systems, such as in vitro cell lines, compared to unperturbed references. Researchers use controlled perturbations in relevant biological model systems to establish causal relationships between molecular mechanisms, genes, chemical compounds, and disease phenotypes. This causal understanding of foundational biological relationships has the potential to positively impact numerous important societal goals \cite{biermann2017global}, including the production of climate-friendly foods and materials and the development of novel therapeutics that address unmet health needs.

The path to understanding complex biological systems and developing targeted therapeutics hinges on unraveling how cells respond to perturbations.
High-throughput experiments have generated an unprecedented volume of perturbation data spanning thousands of perturbations across diverse readout modalities and biological contexts, from single-cell to in-vivo settings~\cite{shalem2015high, rauscher2016genomecrispr, subramanian2017next, oughtred2019biogrid, replogle2022mapping}.
However, these experiments, while rich in indispensable information, vary dramatically in their protocols, readouts, and model systems, often with minimal overlap.
The vast scale and heterogeneity of this data, compounded by context-specific effects, make it extremely challenging to derive generalizable biological insights that drive scientific discovery\footnote{For example, the relationship between observed PSMA1 gene expression levels across multiple cell lines under STAT1 CRISPRi remains unclear, as it is difficult to disentangle the effects of experimental context from those of the perturbation itself.}.

This fundamental challenge of extracting meaningful biological insights from perturbation data has spurred the development of diverse computational approaches.
Most existing approaches focus specifically on predicting the effects of unobserved perturbations~\citep{kamimoto2023dissecting,roohani2023predicting,yuan2021cellbox,lotfollahi2019scgen,hetzel2022predicting, lotfollahi2023predicting, wu2022predicting, bunne2023learning}.
This addresses a fundamental limitation of experimental methods: it is physically impossible to perform all possible configurations of perturbation experiments due to the effectively infinite number of potential experimental designs\footnote{Considering the time of measurement can be arbitrarily long, the number of experiments that may be conducted is already unbounded based on this dimension alone.}.
For example, GEARS~\cite{roohani2023predicting} leverages gene representations based on domain knowledge to predict the effects of unseen genetic perturbations while also providing a means of identifying genetic interaction subtypes.
CPA~\cite{lotfollahi2023predicting} predicts the effects of unseen perturbation combinations, including drugs as perturbagens and their dosages.
Beyond perturbation effect prediction, some methods focus on other critical biological discovery tasks, such as estimating gene-gene relationships~\citep{chevalley2022causalbench}, learning transferable cell representations~\citep{lopez2022learning, rosen2023universal}, modeling relationships among different types of readouts \cite{schmauch2020deep,arslan2022large,mehrizi2023multiomics}, or aiding experimental design~\citep{mehrjou2021genedisco,lyle2023discobax}.

More recently, foundation models~\citep{theodoris2023transfer,cui2023scgpt,hao2023large,chen2023genept} have emerged that are pre-trained on large collections of transcriptomics data to address multiple biological discovery tasks through task-specific fine-tuning pipelines.
\rnn{These models, exemplified by Geneformer~\citep{theodoris2023transfer} and scGPT~\citep{cui2023scgpt}, employ Transformer-based encoders to infer gene and cell representations from gene expression measurements.
While their encoder-based approach offers a compelling advantage — the ability to make predictions for previously unseen contexts by extracting contextual information from gene expression profiles — it faces two significant limitations.
First, the low signal-to-noise ratio in high-throughput screens can pose a challenge to the encoder's ability to extract reliable contextual information, which may result in limited prediction performance.
Second, these models are primarily designed for transcriptomics data and are not inherently structured to accommodate diverse perturbation experiments that use other perturbation and readout modalities, such as for example chemical perturbations or low-dimensional screens measuring cell viability.}

\rnn{To enable in-silico biological discovery from a diverse pool of perturbation experiments, we demonstrate that heterogeneous experimental data, regardless of perturbation type or readout modality, can be integrated into a Large Perturbation Model (\themethod{}) by representing perturbation, readout, and context as disentangled dimensions.}
Similar to foundation models~\citep{theodoris2023transfer,cui2023scgpt}, \themethod{} is designed to support multiple biological discovery tasks, including perturbation effect prediction, molecular mechanism identification, and gene interaction modeling.
\themethod{} is trained to predict outcomes of novel \emph{in-vocabulary} combinations of perturbations, contexts, and readouts.
\themethod{} introduces two architectural innovations that support its primary goal of handling heterogeneity in perturbation data.
First, \themethod{} disentangles the dimensions of perturbation (P), readout (R), and context (C), representing each dimension as a separate conditioning variable.
Second, \themethod{} adopts a decoder-only architecture, meaning it does not explicitly encode observations or covariates.
The PRC-disentangled, encoder-free \themethod{} architecture introduces key advantages:

\begin{itemize}
    \item \emph{Seamless integration of diverse perturbation data.} By representing perturbation experiments as P-R-C dimensions, \themethod{} effectively learns from heterogeneous experiment data across diverse readouts (e.g., transcriptomics, viability), perturbations (CRISPR, chemical), and experimental contexts (single-cell, bulk) without loss of generality and regardless of dataset shape or format.
    \item \rnn{\emph{Contextual representation without encoder constraints.} Encoder-based models assume that all relevant contextual information can be extracted from observations and covariates which may be limiting due to high variability in measurement scales across contexts and a potentially low signal-to-noise ratio.
    In contrast, \themethod{} learns perturbation-response rules disentangled from the specifics of the context in which the readouts were observed. A limitation of this approach is the inability to predict perturbation effects for \emph{out-of-vocabulary} contexts.}
    \item \emph{Enhanced predictive accuracy across experimental settings.} By leveraging its PRC-disentangled architecture and decoder-only design, \themethod{} consistently achieves state-of-the-art predictive accuracy across experimental conditions. 
\end{itemize}

When trained on a pool of experiments, we demonstrate experimentally that \themethod{} achieves state-of-the-art performance in post-perturbation outcome prediction.
Additionally, \themethod{} provides meaningful insights into the molecular mechanisms underlying perturbations, readouts, and contexts.
\themethod{} enables the study of drug-target interactions for chemical and genetic perturbations in a unified latent space, accurately associates genetic perturbations with functional mechanisms, and facilitates the inference of causal gene-to-gene interaction networks.
To demonstrate the potential of \themethod{} for therapeutic discovery, we used a trained \themethod{} to identify potential therapeutics for autosomal dominant polycystic kidney disease (ADPKD).
Finally, we show that the superior performance of \themethod{} compared to existing methods is driven by its ability to leverage perturbation data at scale, achieving significantly improved performance as more data becomes available for training.

\newpage

\begin{figure*}[t!]
\centering
\includegraphics[width=0.98\textwidth]{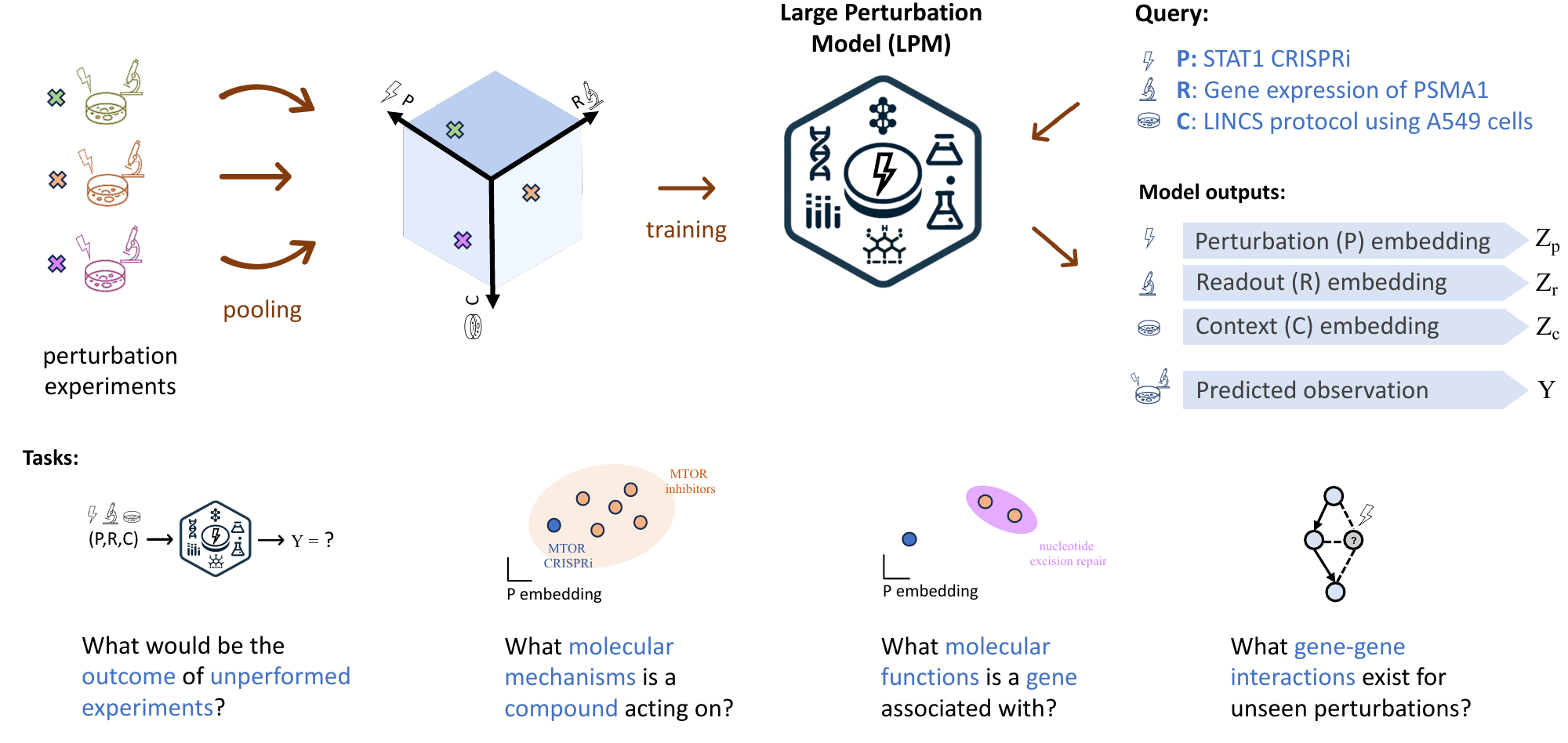}
\caption{\textbf{Large Perturbation Model (\themethod{}) integrates data from multiple perturbation experiments to address a range of biological discovery tasks.} Perturbation experiments originating from different studies are pooled together. Each experiment is placed in the space spanned by perturbations (P), readouts (R) and experimental contexts (C), where multiple experiments generally only partially overlap in the three-dimensional (P,R,C) space (top left). A large perturbation model (\themethod{}; central icon) is trained on pooled perturbation data and can be queried with the symbolic representation of perturbation, readout, and context of experiments of interest to generate embeddings and predict outcomes even for configurations that were not observed during training. \themethod{} embeddings and predictions carry rich information for a range of biological discovery tasks using transfer learning (bottom).}
\label{fig:intro}
\end{figure*}

\section{Results}
\label{sec:results}

Large Perturbation Model (\themethod{}) \rnn{is a deep-learning model that integrates information from pooled perturbation experiments} (\cref{fig:intro}).
We train \themethod{} to predict the outcome of a perturbation experiment based on the symbolic representation of the perturbation, readout, and context (the P,R,C tuple).
\themethod{} features a PRC-conditioned architecture (method details in \Cref{sec:methods}) which enables learning from heterogeneous perturbation experiments that do not necessarily fully overlap in the perturbation, readout, or context dimensions.
By explicitly conditioning on the representation of an experimental context, \themethod{} learns perturbation-response rules disentangled from the specifics of the context in which the readouts were observed.
\themethod{} predicts unseen perturbation outcomes, and its information-rich generalisable embeddings are directly applicable to various other biological discovery tasks (\cref{fig:intro}).
Each task is backed by experimental evidence outlined in the following sections:
\Cref{sec:prediction} covers post-perturbation outcome prediction, \Cref{sec:pharmacogenetic-space} explores pharmacogenetic mechanisms and functions, and
\Cref{sec:genenetworkinference} supports inference of genetic interactions.
Additionally, \Cref{sec:adpkd} further demonstrates the potential of LPM in drug discovery, while \Cref{sec:scaling} investigates the scaling behaviour of \themethod{}.

\subsection{Predicting outcomes of unobserved perturbation experiments}
\label{sec:prediction}

\begin{figure*}[hp]
\centering
    \begin{subfigure}[t]{0.03\textwidth} 
        \textbf{a}
    \end{subfigure}
    \begin{subfigure}[t]{0.96\textwidth}
        {Genetic and chemical perturbations (z-normalized, all readouts, single-cell and bulk)}\hrulefill\linebreak
        \includegraphics[width=\textwidth, valign=t]{figs/performance_benchmark_aggregate_Pearson}
    \end{subfigure}\vspace{1em}
    \begin{subfigure}[t]{0.03\textwidth}
        \textbf{b}
    \end{subfigure}
    \begin{subfigure}[t]{0.4465\textwidth}
        {Genetic perturbations (log-normalized, single-cell, highly variable)}\hrulefill\linebreak
        \includegraphics[width=\textwidth, valign=t]{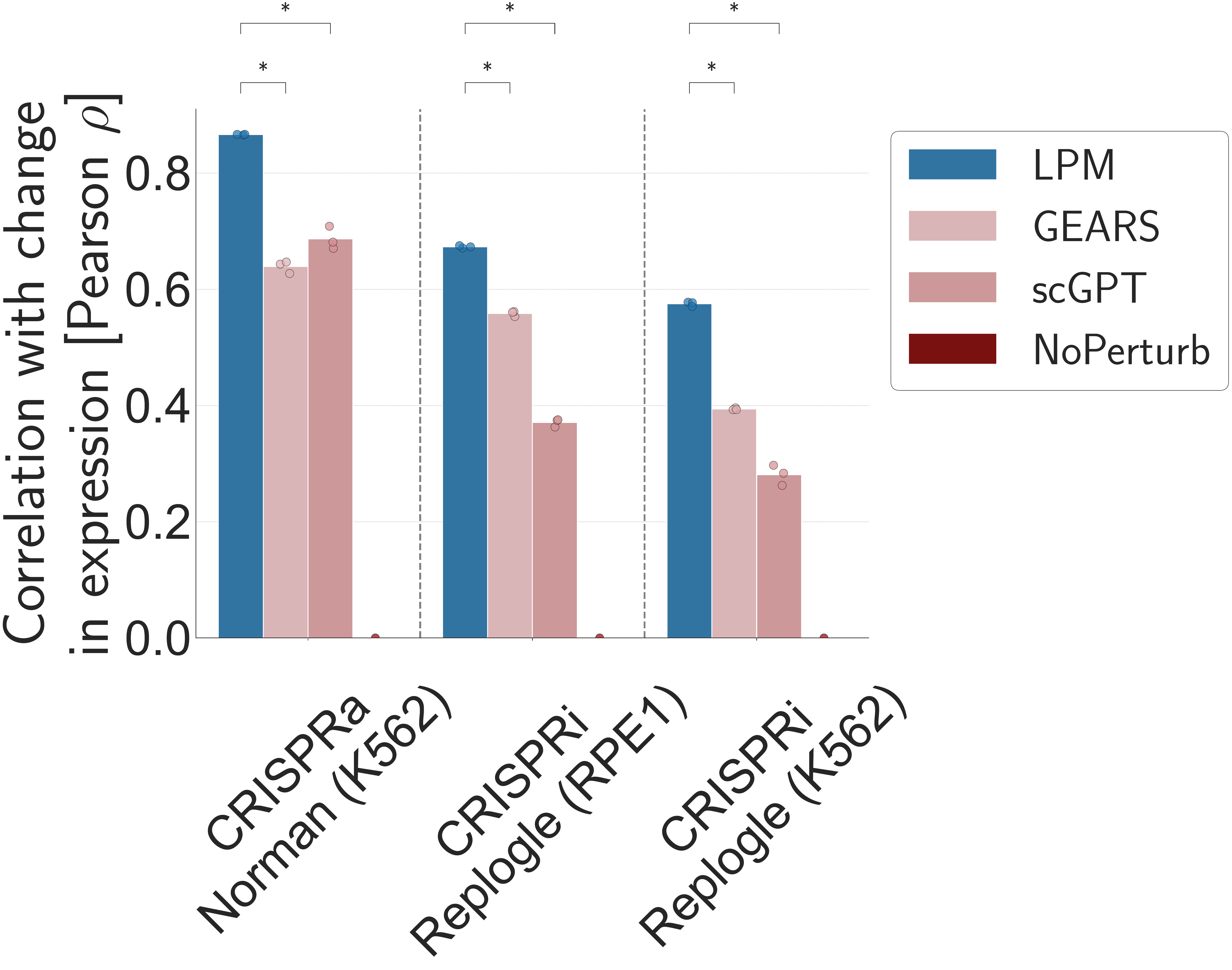}
    \end{subfigure}\hspace{0.85em}
    \begin{subfigure}[t]{0.03\textwidth}
        \textbf{c}
    \end{subfigure}
    \begin{subfigure}[t]{0.4465\textwidth}
        {Genetic perturbations (log-normalized, single-cell, top 20 diff. expressed)}\hrulefill\linebreak
        \includegraphics[width=\textwidth, valign=t]{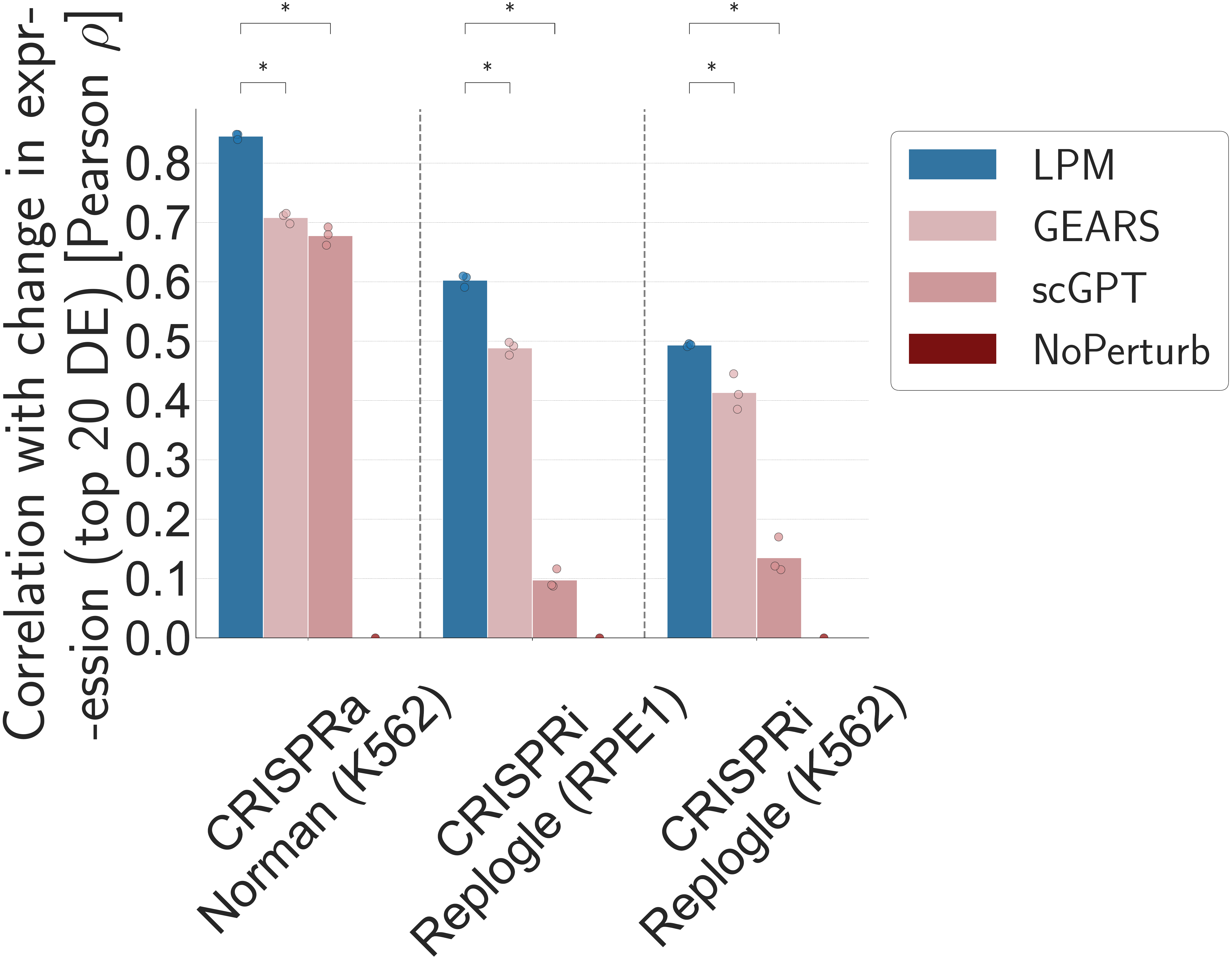}
    \end{subfigure}
\caption{
\textbf{\themethod{} outperforms existing methods in predicting post-perturbation gene expression.}
We compared the performance of \themethod{} to state-of-the-art baselines across a variety of experimental settings, contexts and for different perturbation types. \rnn{
\textbf{a.} A comparison of methods for post-perturbation expression prediction using z-normalized data including all readouts comparing Pearson correlation (y-axis) on held-out test data from eight experimental contexts (x-axis) including single-cell (\citet{replogle2022mapping}), bulk (LINCS \cite{subramanian2017next}), genetic (CRISPRi and CRISPR-KO) and chemical compound interventions. In addition, we performed a comparison methods for post-perturbation expression prediction that replicates the preprocessing methodology from \citet{roohani2023predicting} and \citet{cui2023scgpt}). In this comparison, we calculated Pearson correlation between true and predicted changes in log-normalized expression (control versus perturbed) measured on held-out test data \textbf{b.} for all genes and \textbf{c.} on the subset of the top 20 differentially expressed transcripts (y-axis). \citet{norman2019exploring} includes both single and multi-perturbation data. Across all tested settings, perturbation types and contexts, \themethod{} significantly outperforms state-of-the-art baselines. Bars indicate average performance across different seeds (dots on top of bars). Embedding (\enquote{emb} in parentheses) next to a baseline indicates that we used embeddings that were fine-tuned via a standardized Catboost strategy for evaluation (see \Cref{sec:exp-setup} for details). For baselines without this indication, we used author instructions for generating the post-perturbation expression predictions. Not all methods are suitable for all settings that \themethod{} operates on (e.g., GEARS \cite{roohani2023predicting} requires single-cell resolved data) and are therefore not included in all comparisons}.
Stars indicate statistical significance (one-sided Mann-Whitney, * = $p \le 0.05$).
}
\label{fig:performance-benchmark}
\end{figure*}

We evaluated the performance of {\themethod} in predicting gene expression for unseen perturbations against state-of-the-art baselines, including the compositional perturbation autoencoder (CPA) \cite{lotfollahi2023predicting} and the graph-enhanced gene activation and repression simulator (GEARS) \cite{roohani2023predicting}. We also included baseline models that combined a Catboost regressor \citep{prokhorenkova2018catboost} with existing gene embeddings derived from biological databases (STRING \cite{szklarczyk2019string}, Reactome \cite{fabregat2018reactome} and Gene2Vec \cite{du2019gene2vec}), single-cell foundation models based on pooled gene expression data not under perturbations (Geneformer \cite{theodoris2023transfer} and scGPT \cite{cui2023scgpt}) and natural language descriptions of genes processed through ChatGPT (GenePT \cite{chen2023genept}). \rnn{For scGPT and Geneformer, we either fine-tuned the models according to their respective instructions or used their embeddings with a CatBoost model (indicated as \enquote{emb}).} Additionally, we included the \enquote{NoPerturb} baseline \cite{roohani2023predicting} that assumes that the perturbation does not induce a change in expression. We note that no other baseline model supports predicting experimental results for chemical perturbations and that GEARS, CPA and scGPT (following author instructions) require single cell resolved data.

To evaluate the performance of \themethod{}, we used cross validation and held out a single experimental context as the target context for each fold. Within the target context, test and validation data were randomly held out (stratified by perturbation) and not available for training while the remaining target context data and data from all non-target contexts served as training data for {\themethod} (experimental details provided in \cref{sec:exp-setup}). For GEARS and Catboost-based models, only data from the target context were used since including additional contexts did not benefit those methods. For CPA, due to architectural constraints, we could only include single cell data from the same experimental studies (i.e., all Replogle data). For each target context, we trained models for different random seeds to quantify uncertainty (additional metrics available in \cref{sec:other-results}). 

In order to robustly evaluate the performance of \themethod{}, we conducted a representative array of experiments that covers (i) a range of experimental contexts, (ii) different perturbation types (chemical, genetic), and (iii) varying preprocessing strategies. Across all studied experimental settings, \themethod{} consistently and significantly outperformed the state-of-the-art baselines, regardless of preprocessing methodology (replicating the methodology used in \cite{roohani2023predicting} and \cite{cui2023scgpt} for highly variable and top 20 differentially expressed genes, and also in z-normalised data across all readouts), experimental context and perturbation type under study (\Cref{fig:performance-benchmark}; additional evaluation metrics provided in \Cref{fig:quantitative-analysis}). The datasets include single-cell (\citet{replogle2022mapping}) and bulk (LINCS \cite{subramanian2017next}) data, genetic (CRISPRi,CRISPRa and CRISPR-KO) and chemical compound perturbations, and single and multi-perturbation (\citet{norman2019exploring}) settings. Lastly, our supplementary materials (\cref{fig:viability-experiments}) present data from \citet{horlbeck2018mapping}, which included viability readouts for pairwise CRISPRi perturbations, to demonstrate that \themethod{} is effective even in low-dimensional settings with non-transcriptomic readouts.

\begin{figure*}[tp]
\centering
  \begin{subfigure}[t]{0.03\textwidth}
  \textbf{a}
  \end{subfigure}
  \begin{subfigure}[t]{0.96\textwidth}
	\centering
	\includegraphics[width=0.95\textwidth, valign=t]{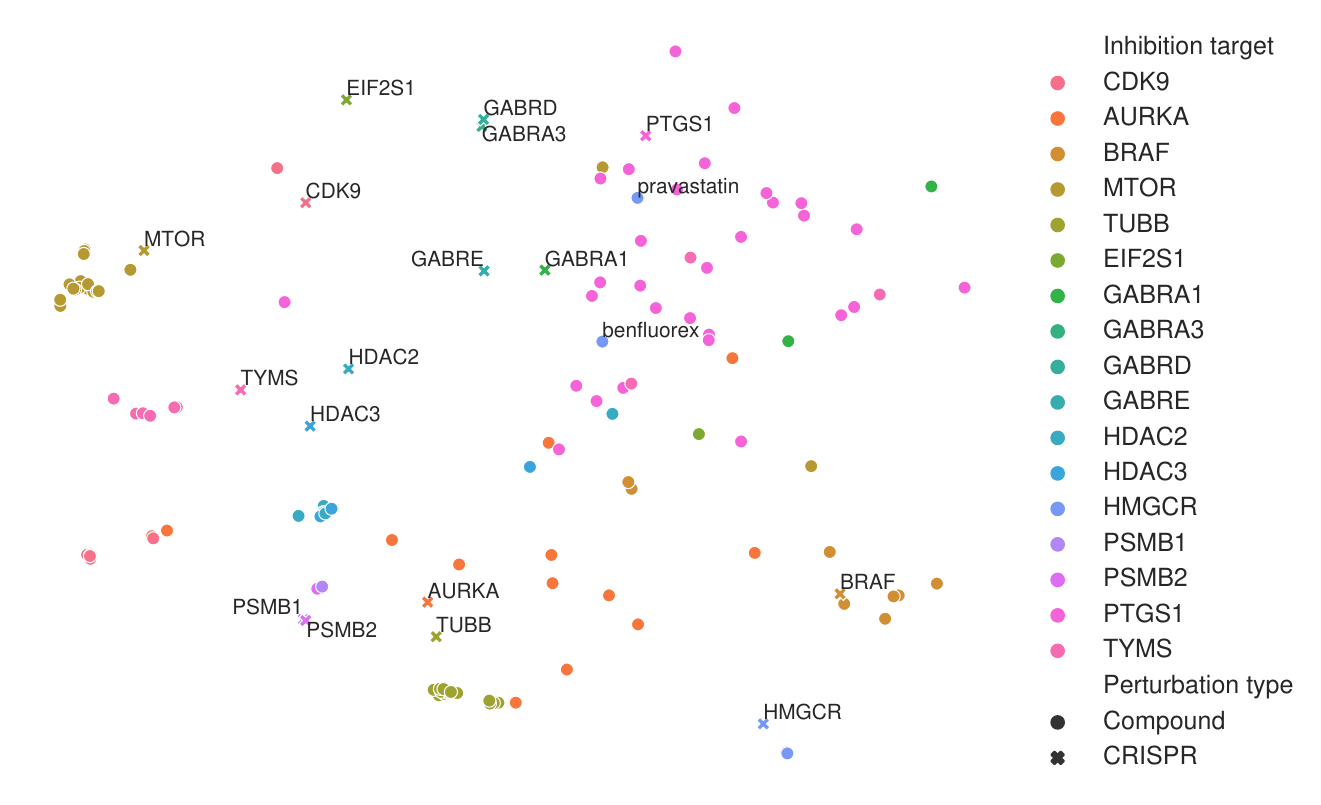}
  \end{subfigure}
  \begin{subfigure}[t]{0.03\textwidth}
  \textbf{b}
  \end{subfigure}
  \begin{subfigure}[t]{0.4\textwidth}
      \vspace{0.1em}
      \centering
      {\small
      \begin{tabular}{lr}
        \toprule
        Compound & RMSE distance \\
         & to HMGCR \\
        \midrule
        pitavastatin & 0.211 \\
        simvastatin & 0.229 \\
        lovastatin & 0.236 \\
        atorvastatin & 0.237 \\
        rosuvastatin & 0.239 \\
        cerivastatin & 0.239 \\
        fluvastatin & 0.242 \\
        mevastatin & 0.248 \\
        \rowcolor{lightgray}
        benfluorex & 0.282 \\
        \rowcolor{lightgray} pravastatin & 0.296 \\
      \bottomrule
    \end{tabular}}
  \end{subfigure}
  \begin{subfigure}[t]{0.03\textwidth}
    \textbf{c}
  \end{subfigure}
  \begin{subfigure}[t]{0.51\textwidth}
        \par\smallskip
	\includegraphics[width=\textwidth, valign=t]{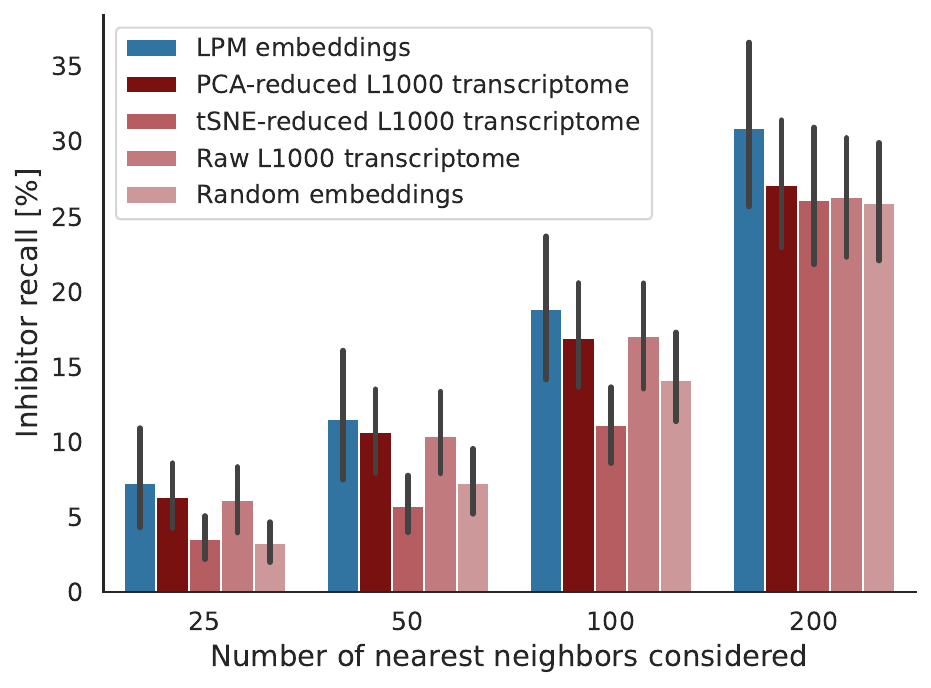}
  \end{subfigure}\hfill
\caption{\textbf{\themethod{} learns a joint space of compound and CRISPR perturbations.} \textbf{a.}
The latent space of compound and CRISPR knockouts (reduced to two-dimensions via t-SNE) reflects known groupings of compound and genetic perturbations that target the same molecular mechanisms in bulk LINCS L1000 data from \cite{subramanian2017next}. Genes targeted by corresponding CRISPR and compound inhibitors are color-coded in matching colors. \textbf{b.} Root mean squared error (RMSE) distances of known HMGCR inhibitors (statins) to the corresponding CRISPR-HMGCR perturbation in the embedding space of the \themethod{}. Two outliers are highlighted in grey and additionally annotated in sub-figure a: {benfluorex} (withdrawn for cardiovascular side effects \cite{tribouilloy2011benfluorex}) and {pravastatin} (shown to have low correlation to other statins \cite{jiang2021control} and additional anti-inflammatory effects \cite{blake2000statins,mcgown2010beneficial,sommeijer2004anti}). \textbf{c.}  Using the distance between \themethod{} perturbation embeddings for chemical and genetic perturbations achieves higher recall of known inhibitors of the respective genetic target than the embeddings derived from post-perturbation L1000 transcriptome profiles. }
\label{fig:lincs-embeddings}
\end{figure*}

\subsection{Learning a joint perturbation space between compound and CRISPR perturbations}

To evaluate the ability of \themethod{} to support the generation of insights across different types of perturbations, we trained an instance of \themethod{} using all available data from LINCS experiments \cite{subramanian2017next} involving both genetic and pharmacological perturbations across a total of 25 experimental contexts with unique combinations of cellular contexts and perturbation types. To our knowledge, this is the first computational model to integrate genetic and pharmacological perturbations within the same latent space, enabling novel approaches for studying drug-target interactions. When studying t-distributed stochastic neighbour embeddings (t-SNE) \cite{van2008visualizing} of the perturbation embedding space learnt by the \themethod{}, we found that pharmacological inhibitors of molecular targets are consistently clustered in close proximity to genetic CRISPR interventions that target the same genes (\cref{fig:lincs-embeddings}a). For example, genetic perturbations targeting MTOR and compounds inhibiting MTOR and also genetic perturbations from the same pathway, e.g. PSMB1 and PSMB2, or HDAC2 and HDAC3, were clustered closely together. Qualitatively, we found that anomalous compounds that were placed distant from their putative target had been reported to have off target activity (\cref{fig:lincs-embeddings}b), such as benfluorex (withdrawn due to cardiovascular side effects \cite{tribouilloy2011benfluorex}), and pravastatin (shown to elicit expression changes with low correlation to other statins \cite{jiang2023investigating}). Intriguingly, we found that pravastatin moved towards non-steroidal anti-inflammatory drugs (NSAID) that target PTGS1 in the perturbation space (\cref{fig:lincs-embeddings}a) - indicating a potential additional anti-inflammatory mechanism of pravastatin. We found that this movement independently derived by \themethod{} is indeed substantiated by clinical and preclinical observations that ascribe anti-inflammatory effects to pravastatin \cite{blake2000statins,mcgown2010beneficial,sommeijer2004anti}. To further quantitatively validate these findings, we systematically compared known inhibitors of a genetic target with the genetic perturbation in embedding space as a reference. We evaluated the neighborhood of the reference in various embedding spaces and found that perturbation embeddings derived from \themethod{} achieve considerably higher recall of known inhibitors of genetic targets compared to embeddings derived from post-perturbation L1000 transcriptome profiles or dimensionality reduced versions thereof (\cref{fig:lincs-embeddings}c).

\vspace{-0.5cm}
\subsection{Learned embeddings reflect known biological relationships}
\label{sec:pharmacogenetic-space}

\begin{figure*}[tp]
\vspace{-1.5em}
      \centering
      \begin{subfigure}[t]{0.03\textwidth}
        \textbf{a}
      \end{subfigure}
      \begin{subfigure}[t]{0.96\textwidth}
    	\includegraphics[width=\textwidth, valign=t]{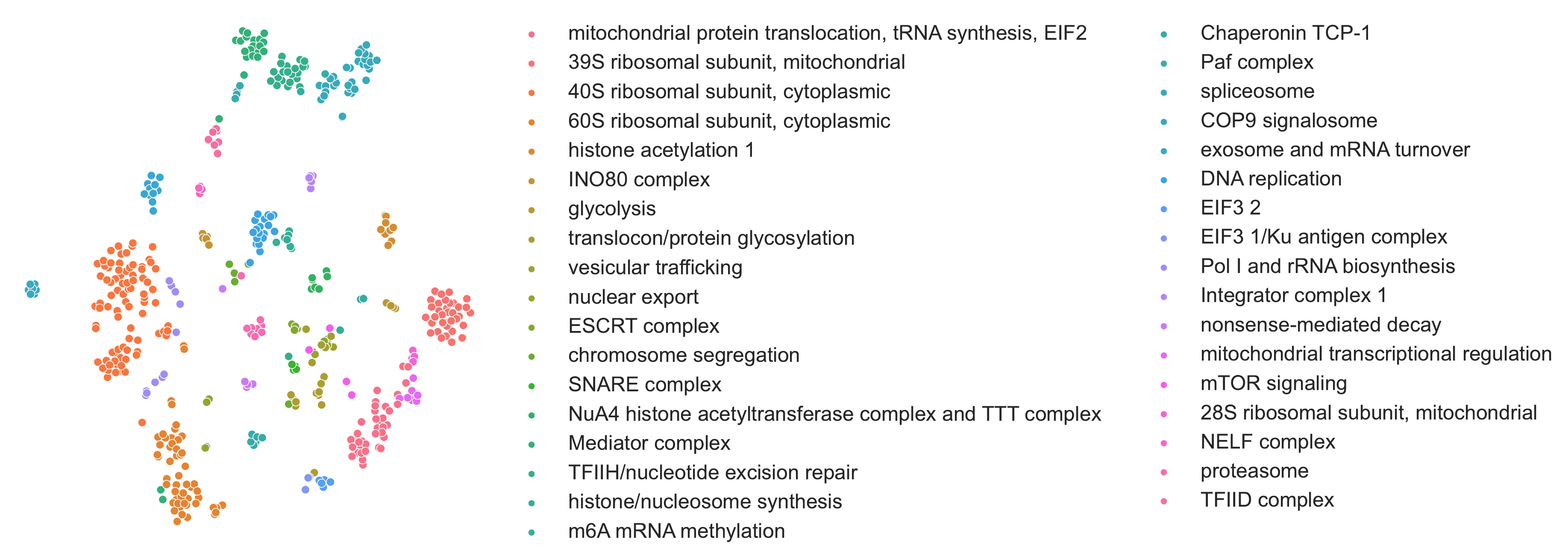}
      \end{subfigure}
      \par\medskip
      \begin{subfigure}[t]{0.03\textwidth}
        \textbf{b}
      \end{subfigure}
      \begin{subfigure}[t]{0.46\textwidth}
            \centering
    	\includegraphics[width=0.8\textwidth, valign=t]{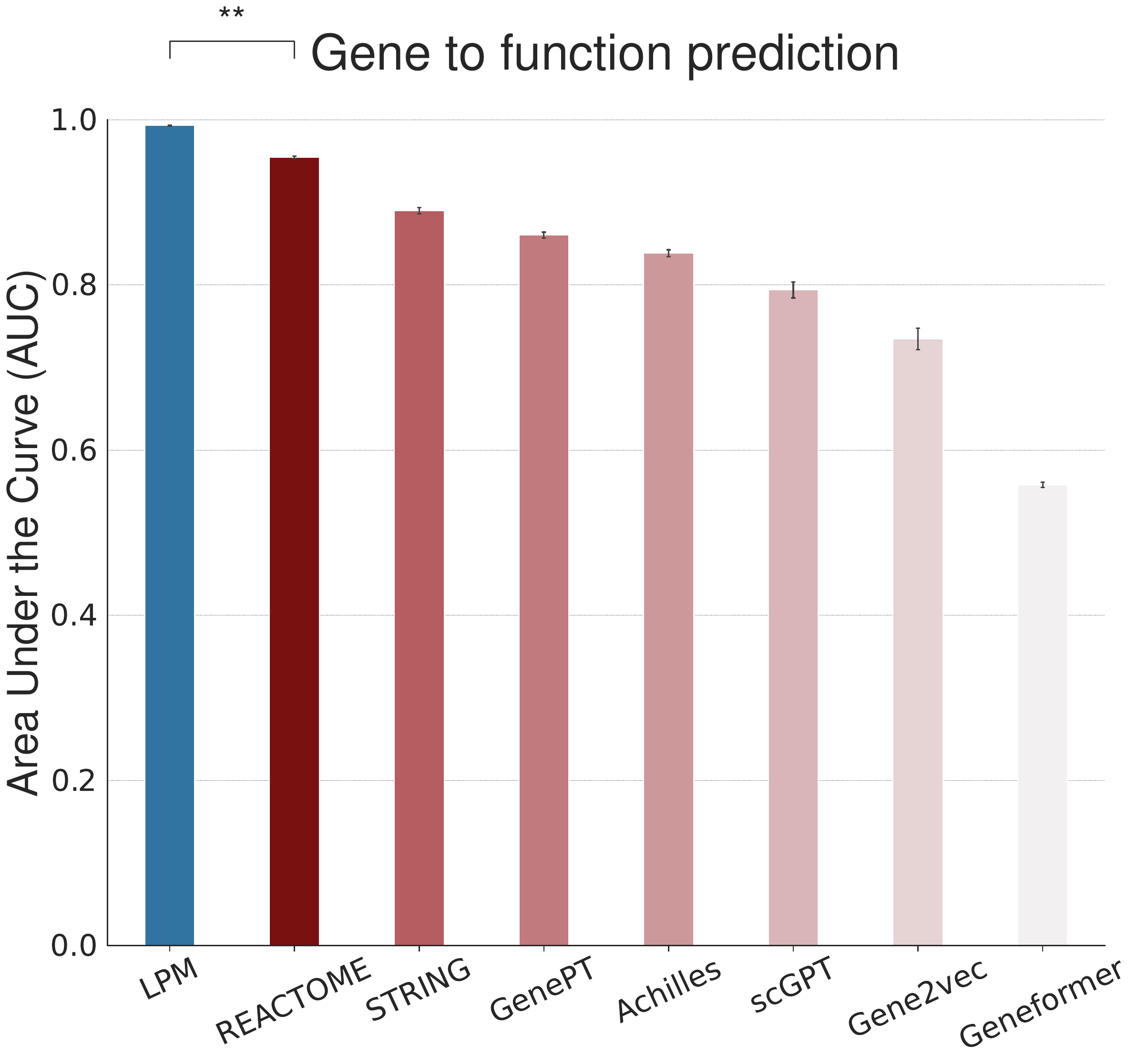}
       \end{subfigure}\hfill
      \begin{subfigure}[t]{0.03\textwidth}
        \textbf{c}
      \end{subfigure}
      \begin{subfigure}[t]{0.46\textwidth}
    	\centering
    	\includegraphics[width=0.95\textwidth, valign=t]{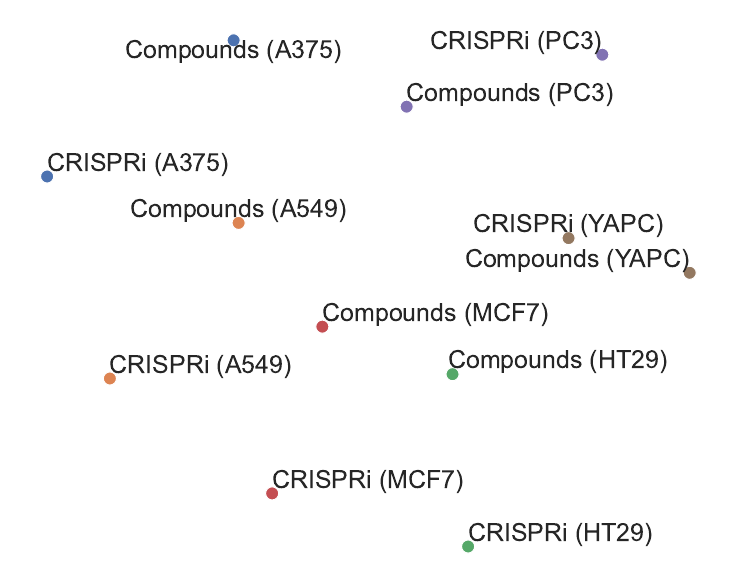}
      \end{subfigure}\hfill
    \caption{\textbf{\themethod{} embeddings reflect rich biological relationships.}
    \textbf{a.} \themethod{} perturbation (P) embeddings (t-SNE embedded in 2D space). Each point represents a CRISPRi perturbation color-coded by molecular function of its respective genetic target from \cite{replogle2022mapping}. \textbf{b.} \themethod{} perturbation (P) embeddings significantly outperform existing state-of-the-art gene embeddings derived from large-scale genetic screens and public pathway and interaction databases in predicting gene function annotations from \cite{replogle2022mapping} (p $\leq0.01$; one-sided Mann-Whitney-Wilcoxon test, 5 random seeds). \textbf{c.} \themethod{} context (C) embeddings (2D t-SNE representation) quantify similarity between experimental contexts. Intriguingly, we found that contexts are grouped with respect to the model system under study (shown in the figure) or by type of perturbation (not shown), depending on the t-SNE random seed used.
}
\label{fig:replogle-embeddings}
\end{figure*}

To evaluate the degree to which \themethod{} perturbation embeddings correspond to known biological functions, we extracted perturbation embeddings for well-characterized perturbations from an {\themethod} trained on pooled single-cell perturbation data \cite{replogle2022mapping} and compared genetic perturbations to gene function annotations as curated by \citet{replogle2022mapping} using the CORUM protein complex \cite{giurgiu2019corum} and STRING \cite{szklarczyk2019string} databases. We found that \themethod{} implicitly organises perturbations according to their molecular functions (\cref{fig:replogle-embeddings}a), and that these embeddings are significantly (p $\leq0.01$) more predictive of gene function annotations than existing state-of-the-art gene perturbation embeddings (\cref{fig:replogle-embeddings}b), including those derived from curated databases such as STRING \cite{szklarczyk2019string} and Reactome \cite{fabregat2018reactome}, derived from co-expression datasets in Gene2Vec \cite{du2019gene2vec} and derived from the single-cell unperturbed gene expression foundation models Geneformer \cite{theodoris2023transfer} and scGPT \cite{cui2023scgpt}) and gene embeddings based on natural language descriptions processed through ChatGPT (GenePT \cite{chen2023genept}).

To qualitatively assess the information contained within context representations of \themethod{}, we employed the \themethod{} model trained on combined LINCS data from the perturbation embedding experiment above to generate context embeddings. We found that - depending on the t-SNE random seeds used - cell types either tend to cluster together with matching cell types from other experiments (\cref{fig:replogle-embeddings}c), or the context embeddings tend to cluster based on the perturbation methodology (CRISPR vs. compound screens; not depicted). The qualitative results imply that the information contained within the learned context embeddings carries information regarding biological semantics and could thus be valuable in downstream analyses, e.g. for quantifying the similarity of contexts.

\begin{figure*}[t!]
\vspace{-2.0em}
\centering
  \begin{subfigure}[t]{0.03\textwidth}
    \textbf{a}
  \end{subfigure}
  \begin{subfigure}[t]{0.33\textwidth}
	\includegraphics[width=\textwidth, valign=t]{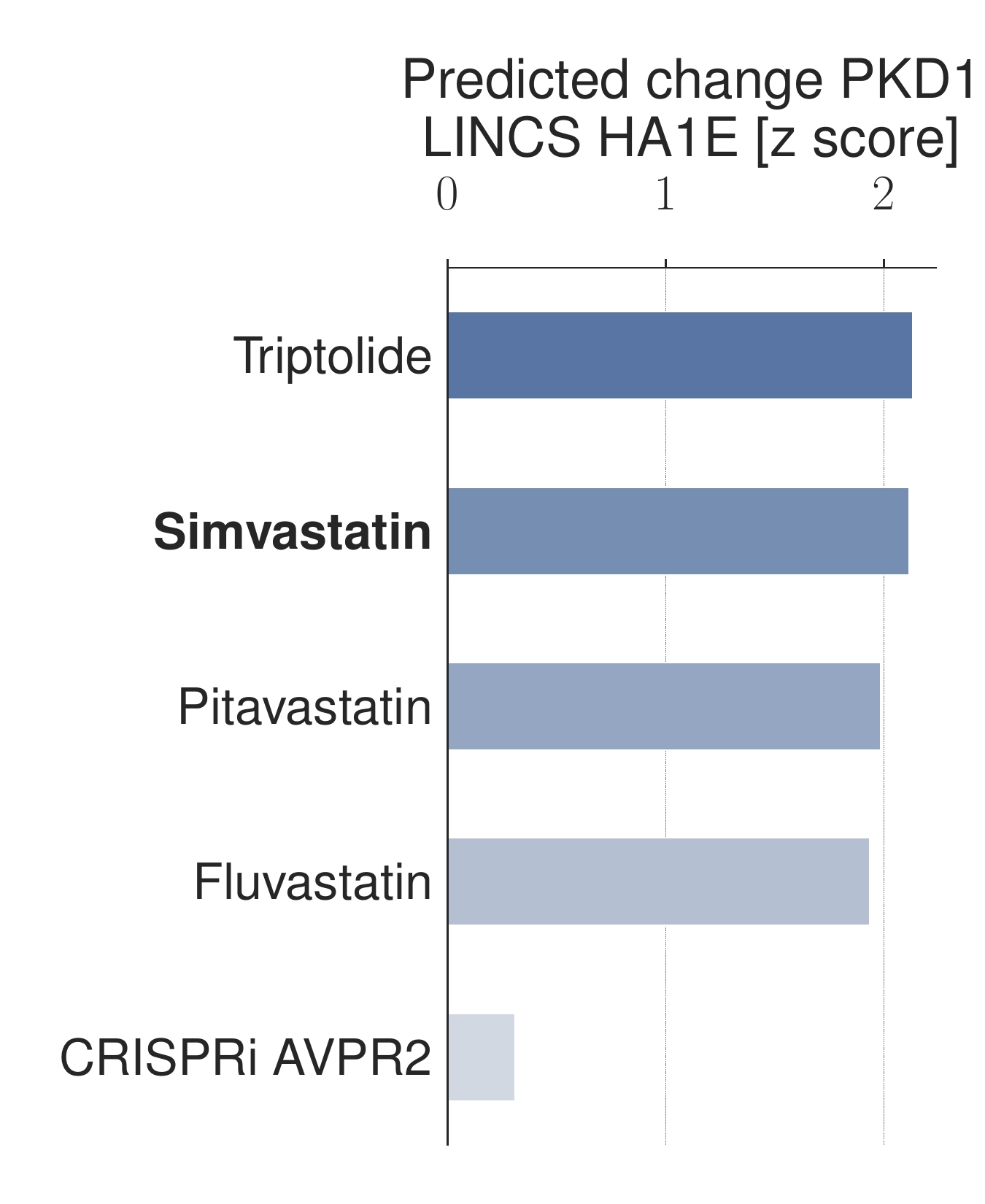}
  \end{subfigure}
  \begin{subfigure}[t]{0.03\textwidth}
    \textbf{b}
  \end{subfigure}
  \begin{subfigure}[t]{0.54\textwidth}
	\includegraphics[width=\textwidth, valign=t]{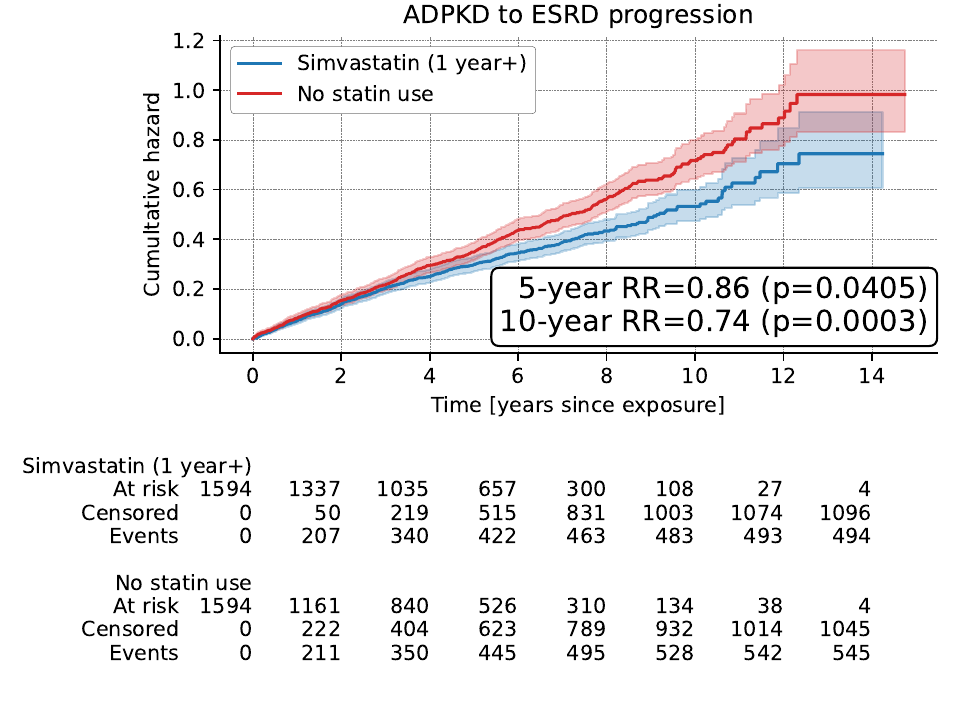}
  \end{subfigure}
  \hfill
\caption{\textbf{In-silico discovery of potential therapeutics for autosomal dominant polycystic kidney disease (ADPKD).} \textbf{a.} Using \themethod{}, we conducted an in-silico perturbation study in which we identified clinical-stage drugs that are predicted to upregulate PKD1 in embryonic kidney cells. A lack of functional copies of PKD1 is hypothesised to be causally involved in ADPKD pathogenesis and progression \cite{hopp2012functional,rossetti2009incompletely,gainullin2015polycystin}. We found that triptolide, simvastatin (bold) and other statins, are the top predicted upregulators of PKD1 among clinical-stage drugs. For reference, we also include the predicted CRISPRi on vasopressin receptor 2 (AVPR2) to simulate the effect of the FDA approved AVPR2 antagonist tolvaptan \cite{torres2012tolvaptan} that is mechanistically distinct \cite{wang2008vasopressin}. \textbf{b.} Since simvastatin is commonly prescribed for cardiovascular indications, we were able to conduct a retrospective cohort study in large-scale electronic health records to further substantiate the potential efficacy of simvastatin in reducing ADPKD progression in the clinic. Most notably, we found that - among individuals diagnosed with ADPKD - 1 year or longer exposure to simvastatin (blue) is associated with a significant (p $\leq0.05$, 5-year relative risk [RR] $= 0.86$ and 10-year RR $= 0.74$) reduction in progression to end stage renal disease (ESRD) compared to those not exposed to statins (red).}
\label{fig:discovery}
\end{figure*}

\subsection{In-silico discovery of potential therapeutics for autosomal dominant polycystic kidney disease}
\label{sec:adpkd}

We hypothesised that the ability of \themethod{} to conduct perturbation experiments in-silico with high accuracy while reflecting underlying biological function could be used to discover potential candidate therapeutics for diseases with known genetic causes, such as autosomal dominant polycystic kidney disease (ADPKD). ADPKD is a genetic disease suspected to be caused by mutations in PKD1 \cite{reeders1985highly} that are reported to lead to a lack of functional PKD1 - eventually manifesting in dose-dependent cystogenesis \cite{hopp2012functional,rossetti2009incompletely,gainullin2015polycystin,lanktree2021insights}. ADPKD affects more than 12 million people worldwide \cite{radhakrishnan2022management} and may lead to severe long-term complications, such as end stage renal disease (ESRD) and the dependence on dialysis or a kidney transplant. There are no curative treatments available for ADPKD. A potential hypothesis for a therapeutic could be to upregulate expression of the functional allele of PKD1 in heterozygous carriers of PKD1 mutations to make up for the non-functional allele, and thereby reach a sufficient level of functional PKD1 that may inhibit further progression of ADPKD. To identify potential therapeutics that could increase PKD1 expression in individuals with ADPKD, we conducted an in-silico perturbation experiment using an \themethod{} trained on pooled LINCS compound and genetic perturbation data to predict which clinical-stage drugs may lead to upregulation in PKD1 levels in HA1E embryonic kidney cells cultured under the LINCS L1000 protocol \cite{duan2016l1000cds2}. We found that triptolide, simvastatin and other statins were among the top clinical-stage drugs predicted to cause increased PKD1 expression in vitro (\cref{fig:discovery}a). Our findings align well with previous literature, where effects of commercially available statins were shown to increase the expression of PKD1 in pancreatic cancer cell line MiaPaCa-2 \citep{gbelcova2017variability}. \rnn{We note that \citet{huang2023long} found no significant change in PKD1 expression in mice exposed to atorvastatin.} Since simvastatin is an FDA-approved medicine that is prescribed preventatively for cardiovascular indications, we conducted a retrospective, matched cohort study to validate the in-silico hypothesis that simvastatin may lead to reduction in ESRD progression in real-world clinical data from the Optum\textsuperscript{\textcopyright} de-identified Electronic Health Record (EHR) database. Notably, we found that - among individuals diagnosed with ADPKD - exposure to simvastatin over 1 year or longer was associated with a significant decrease (5-year relative risk $= 0.86$, p $= 0.0405$ and 10-year relative risk $= 0.74$, p $= 0.0003$) in progression to ESRD compared to those not exposed to any statins predicted by \themethod{} to increase expression of PDK1 (\cref{fig:discovery}b). Several of the therapeutics predicted to increase PKD1 are substantiated by literature, e.g. pravastatin was shown to be associated with improved kidney markers in a clinical study in young individuals \cite{cadnapaphornchai2014effect} and triptolide led to a reduction of cystogenesis in murine models \cite{leuenroth2007triptolide,leuenroth2008triptolide}. PKD1 was neither measured nor perturbed in LINCS, nor were all \numprint{5310} chemical perturbations tested in HA1E cells, and the in-silico \themethod{} experiments were therefore essential to enable this study. We note that these findings should not be considered definitive and that further research is required to validate and support them.

\subsection{Facilitating inference of causal gene-gene relationships}
\label{sec:genenetworkinference}

\begin{figure*}[!t]
\vspace{-1.5em}
\centering
  \begin{subfigure}[t]{0.03\textwidth}
    \textbf{a}
  \end{subfigure}
  \begin{subfigure}[t]{0.44\textwidth}
	\includegraphics[trim=0.2cm 0.1cm 0.7cm 0cm, clip, width=\textwidth, valign=t]{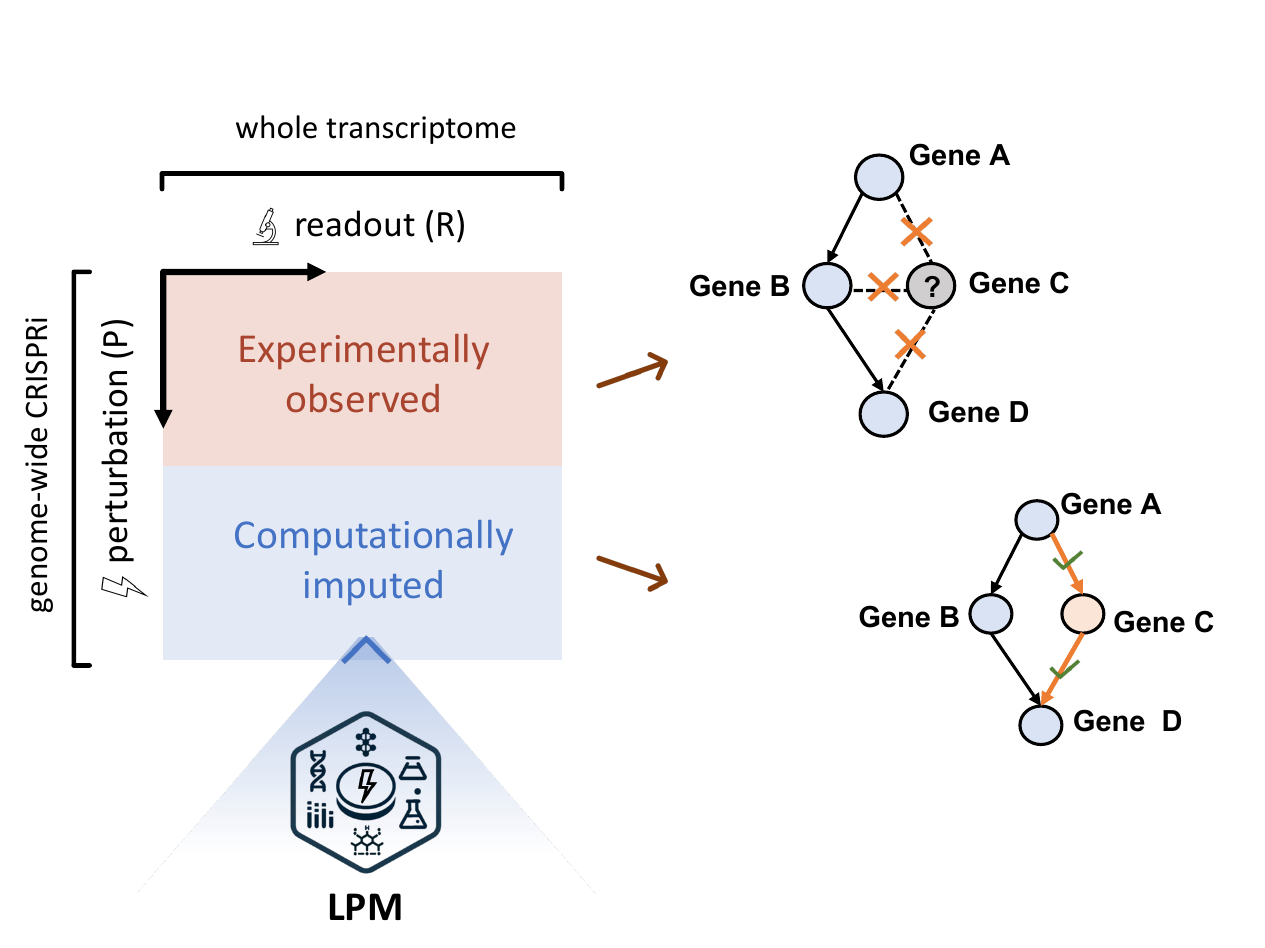}
  \end{subfigure}
  \begin{subfigure}[t]{0.03\textwidth}
    \textbf{b}
  \end{subfigure}
  \begin{subfigure}[t]{0.44\textwidth}
	\includegraphics[width=\textwidth, valign=t]{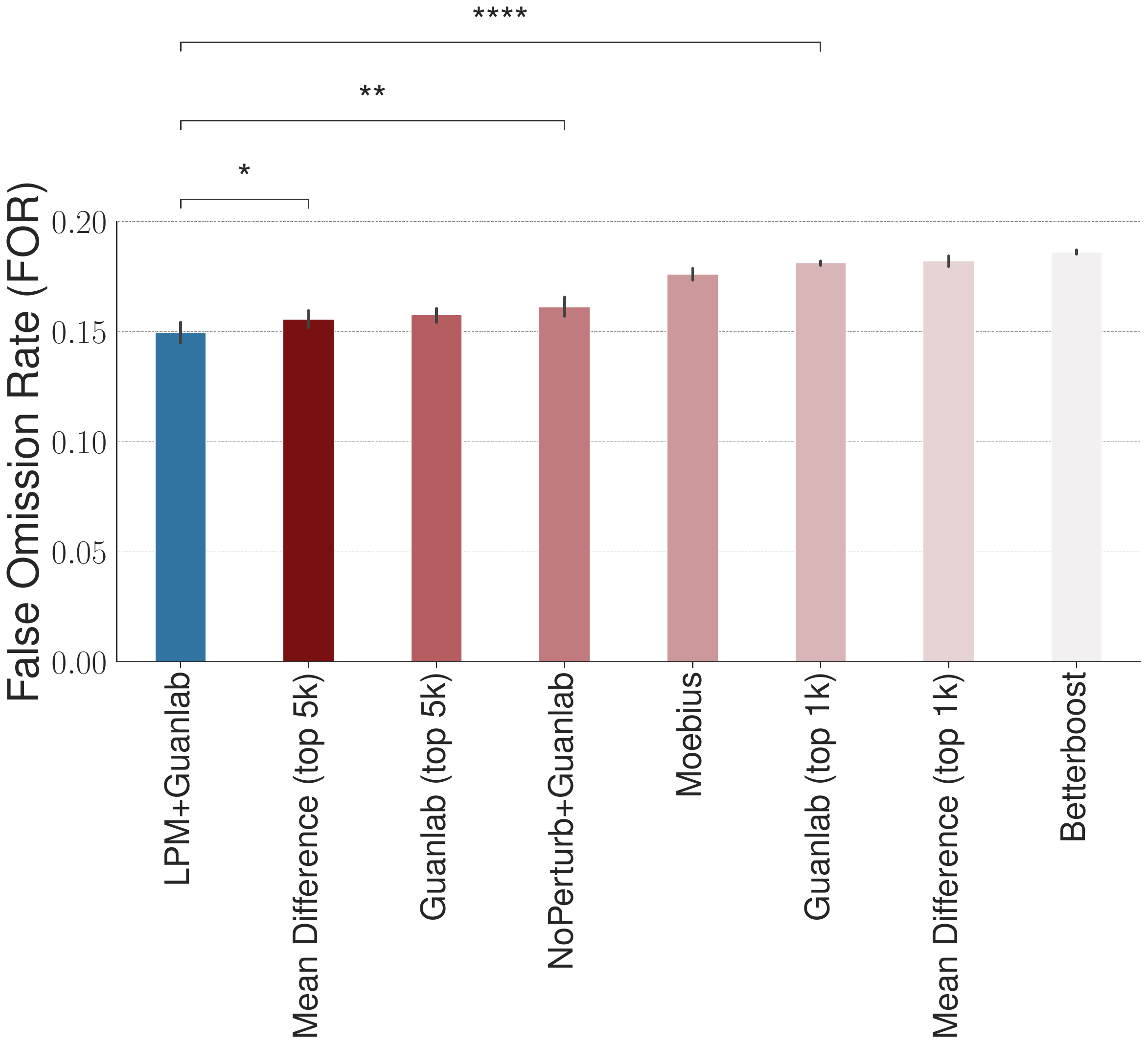}
  \end{subfigure}
\caption{\textbf{\themethod{} improves gene-gene network inference.} \textbf{a.} We used \themethod{} to predict post-perturbation transcriptomes for unseen perturbations, completing a partially observed experimental space where only half (orange area) of the possible CRISPRi perturbations across the genome were experimentally observed. We hypothesised that access to the computationally completed dataset (blue plus orange area) may enable the state-of-the-art Guanlab \cite{deng2023supervised} gene network inference method to more accurately infer the gene-gene interactions for genes not experimentally perturbed. \textbf{b.} We found that the combination of LPM imputation and the Guanlab method (\themethod{}+Guanlab) significantly outperformed existing methods for gene-gene network inference using the partially observed dataset alone in terms of False Omission Rate (FOR) in a gene network inference benchmark \cite{chevalley2022causalbench} using single-cell data from \citet{replogle2022mapping}. Stars indicate statistically significant differences (one-sided Mann-Whitney-Wilcoxon, * = p $\leq0.05$, ** = p $\leq0.01$).}
\label{fig:causality}
\end{figure*}

To assess to what degree the accuracy of the predictions of \themethod{} translate to capturing mechanistic interactions between genes, we used \themethod{} in the context of causal inference of gene interaction networks. Normally, these networks are inferred from perturbation experiments in which only a subset of all genes were perturbed. In contrast, we measured the enhancement in performance when those networks were inferred from the same experimental data enriched with missing, unmeasured CRISPRi perturbations predicted in-silico using \themethod{}. In particular, to perform network inference, we applied corresponding methods that demonstrated best-in-class performance on the recent CausalBench challenge~\citep{chevalley2022causalbench} and were designed specifically for inferring gene-gene networks from perturbational scRNAseq data. We found that augmenting the original data with in-silico perturbation outcomes, prior to applying network inference using above mentioned methods, lead to a significant improvement in terms of False Omission Rate (FOR) in comparison to existing state-of-the-art methods for gene-gene network inference that do not have access to perturbation imputation (\cref{fig:causality}). These results underscore the utility of \themethod{} in supporting the inference of more comprehensive and accurate causal interactions tailored to a given experimental context and the ability of \themethod{} to learn generalisable, causal interactions between perturbations.

\subsection{\themethod{} benefits from more experiments and perturbations}
\label{sec:scaling}

\begin{figure*}[t!]
\vspace{-1em}
\centering
  \begin{subfigure}[t]{0.03\textwidth}
    \textbf{a}
  \end{subfigure}
  \begin{subfigure}[t]{0.44\textwidth}
	\includegraphics[width=\textwidth, valign=t]{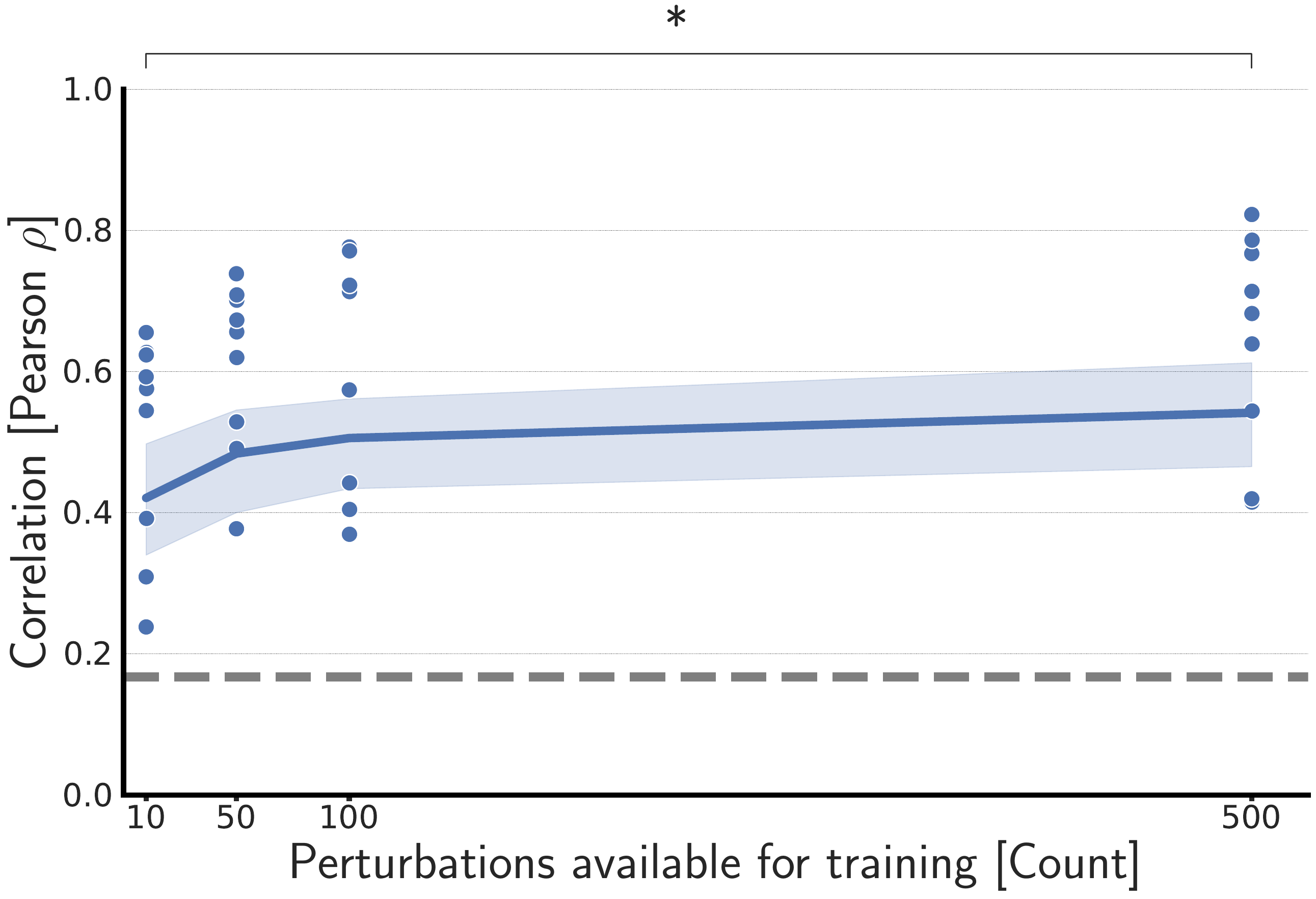}
  \end{subfigure}
  \begin{subfigure}[t]{0.03\textwidth}
    \textbf{b}
  \end{subfigure}
  \begin{subfigure}[t]{0.44\textwidth}
	\includegraphics[width=\textwidth, valign=t]{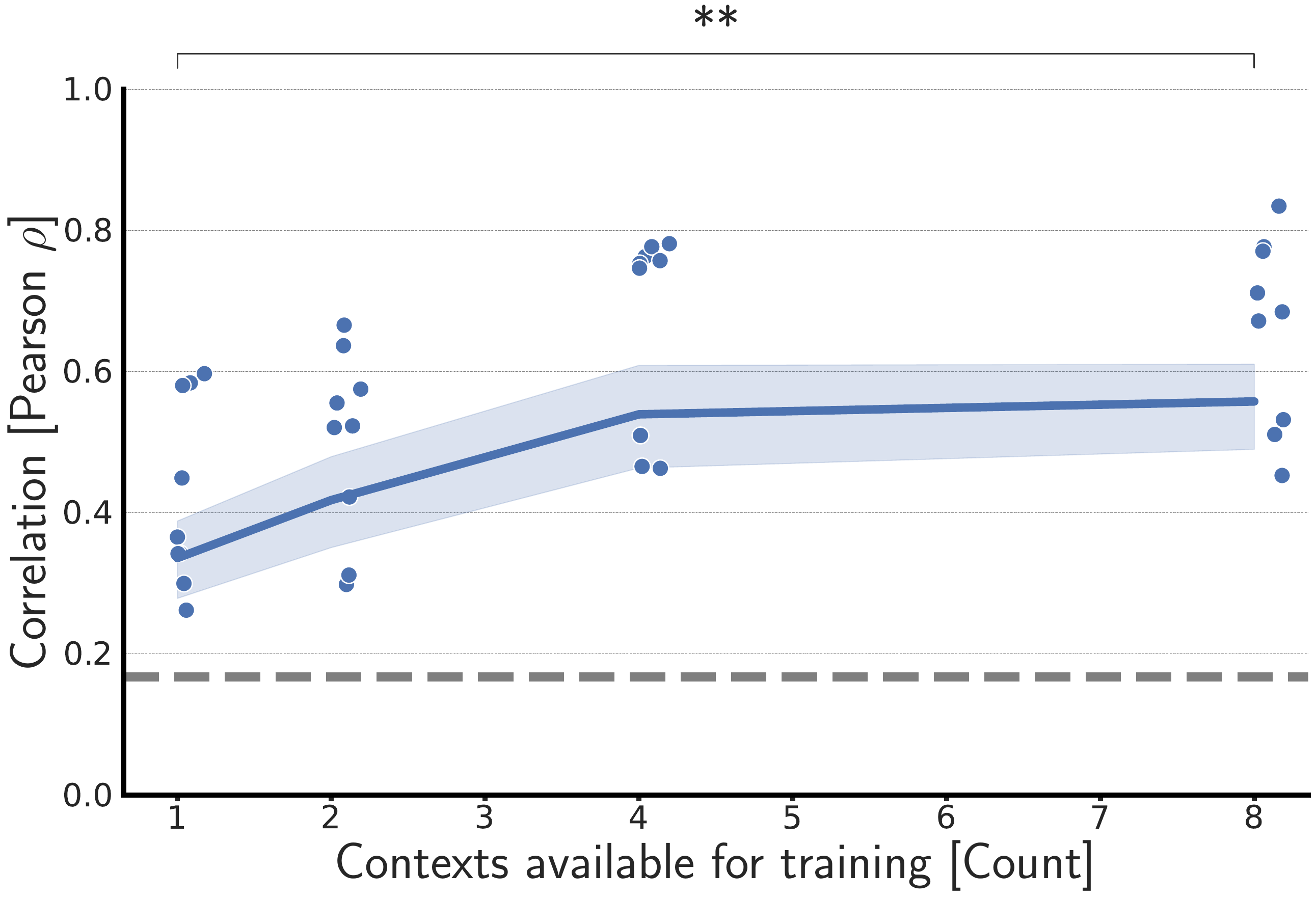}
  \end{subfigure}
  \hfill
\caption{\textbf{The performance of \themethod{} increases as more data is available for training.} Performance comparison in terms of Pearson correlation coefficient $\rho$ in predicting the outcomes of unseen experiments of \themethod{} when varying \textbf{a.} the number of perturbations available for training in a target context and \textbf{b.} the number of different contexts available for training. Dots correspond to individual runs with a different random seed, and the blue line corresponds to the inferred trend. The dashed grey line denotes the performance of the \enquote{NoPerturb} baseline, which does not take perturbation information into account. The performance of \themethod{} increases significantly (p $\leq$ 0.05) when more perturbations and more contexts are available.}
\label{fig:scaling}
\end{figure*}

In contrast to data-rich domains like natural language processing (NLP) where scaling of model performance with additional data has been established experimentally \cite{kaplan2020scaling,hoffmann2022training}, it is not yet clear to what degree in-silico biological discovery can benefit from availability of additional data both across contexts and perturbations for pooling. Establishing data scaling patterns in biology has historically been more difficult than in predominantly digital domains like NLP and computer vision because biological perturbation data can often not be naively aggregated due to the intricate connection between experimental context, data processing methodologies and batch effects. To elucidate the potential performance benefits of additional data for \themethod{}, we computationally evaluated the prediction performance in terms of Pearson correlation coefficient $\rho$ for predicting unseen perturbations when varying the number of datasets covering multiple contexts and perturbations in a single context available for model training (\cref{fig:scaling}). The performance of \themethod{} significantly (p $\leq0.05$) improves both when more datasets covering multiple contexts and when more perturbations in a single context are available for training.

\section{Discussion}

We present the Large Perturbation Model (\themethod{}) \rnn{that integrates data from multiple perturbation experiments to perform biological discovery tasks in silico, including predicting the outcome of unseen perturbation experiments, understanding the shared molecular mechanism of action of chemical and genetic perturbations, and deriving gene-gene interaction networks.} Experimentally, we found that the use of \themethod{} - either independently or in combination with a causal network inference algorithm - significantly outperforms existing state-of-the-art methods, providing an experimental proof of concept for the potential to accelerate biological discovery with computationally generated evidence. The ability to generate unobserved experimental data for critical biological questions, such as what the estimated effects of unseen perturbations would be, could potentially accelerate the generation of insights and could complement experimentally generated data - in particular, in the settings that are difficult, time or resource-intensive to study in real world laboratory experiments. Notably, we found that \themethod{} implicitly learns rich latent space embeddings for perturbations, readouts and experimental contexts as is required to achieve their explicit training objective to predict yet unseen experimental outcomes. The rich latent space embeddings of \themethod{} enables a range of downstream biological discovery tasks (only a subset of the potential use cases are investigated in this study) which demonstrates the versatility and multitask capability of \themethod{} that captures underlying mechanistic relationships in data.

While scaling properties for machine learning models have previously been established in data-rich domains \cite{kaplan2020scaling,hoffmann2022training}, they have to date remained elusive in biology where different formats of datasets, batch effects \cite{leek2010tackling}, experimental variance and poor repeatability \cite{errington2021investigating} make pooling of data across experiments difficult. The observed performance scaling behaviour of \themethod{} on both perturbations and datasets available for training shows that there are integrative modelling approaches that enable effective transfer learning across datasets in biology, and may provide a fertile ground for \themethod{} with an even broader and more comprehensive understanding of mechanisms underlying perturbations, readouts and contexts as more perturbation data becomes available in the future. We note that \themethod{} is not large in terms of parameter counts relative to domains with more readily available data under the current curricula available for training, but the observed scaling behaviour indicates that \themethod{} may - once sufficiently sized data is available - potentially benefit from scale in terms of performance.

One limitation of \themethod{} is its inability to extrapolate to unseen contexts in a zero-shot manner. In fact, \themethod{} can be seen as addressing a subset of tasks compared to related approaches with zero-shot context transfer capabilities (CPA~\citep{lotfollahi2023predicting}, scGPT~\citep{cui2023scgpt}, Geneformer~\citep{theodoris2023transfer}), but offering better performance in this set of tasks. For instance, \themethod{} is particularly useful if there is an existing pool of perturbation screens and the goal is to extract shared biological patterns across those screens, or when interpolation to unseen combinations of perturbations, readouts, and contexts is required. \rnn{The recent trend indicates that in the near future, as perturbational experimental data becomes more abundant, the experimental space will be sufficiently covered, rendering in-vocabulary approaches sufficient for most tasks. Furthermore, there is also potential to use predefined context or perturbation embeddings (as done in GEARS), to facilitate out-of-vocabulary predictions.}

While \themethod{} addresses a key challenge in aggregating data from pooled perturbation experiments, many challenges - including the appropriate curation, standardization and preprocessing - are equally or more important in \themethod{} as they are with other machine learning models. \themethod{} is flexible with respect to the format of data generated, but they must still be meaningful and comparable to each other in terms of data standardisation for \themethod{} to derive a meaningful latent space. The design of \themethod{} makes it flexible to include more structured perturbation descriptions (e.g., including drug dosage \cite{schwab2020doseresponse}), but also a more structured description of readouts (e.g., including the temporal dimension). While the experiments conducted in this study are limited to a restricted set of experimental contexts covered due to the sparsity of diverse publicly available high-throughput perturbation datasets, we expect that including additional training data would further improve performance and coverage of the space of perturbations, readouts and contexts.
\rnn{One additional aspect that should be explored in the future work is training and benchmarking on non-immortalized cell lines. Our study is limited to immortilized cell lines and it is critical that in the future genetically natural cells are used for benchmarking.}
In addition, while the in-silico study of PKD1 upregulators to address ADPKD progression yielded notable qualitative evidence for the utility of \themethod{} in therapeutic discovery, we note that the presented real-world data study was retrospective, and therefore cannot address unobserved confounding. Randomised controlled trials (RCTs) are necessary to conclusively establish the effectiveness of in-silico predicted candidate therapeutics. As a further limitation, we considered only a single genetically validated marker in our ADPKD study. It is important to that additional clinical validation is needed to conclusively prove causality for LPM's predictions in addressing ADPKD. Furthermore, therapeutic candidates must be optimised w.r.t. multiple criteria, including safety, pharmacokinetics and -dynamics.
Finally, we would like to emphasize that (gene-gene) network inference is a distinct and complex field of research, and future studies will need to explore additional datasets and benchmarks to further validate findings in this area. Our study, however, is focused on demonstrating the potency of high-quality perturbation-effect predictors, such as \themethod{}, to complement existing network inference methods.

It would be immensely important for future studies to test LPM on the growing body of high-dimensional omics data derived from clinical settings. This could present valuable opportunities to identify novel therapies or patient cohorts that are likely to respond to specific treatments, thus advancing the field of personalized medicine. By leveraging these datasets, LPM could help pinpoint biomarkers of response (e.g. clinical covariates) and further optimize therapeutic strategies for patients based on their unique molecular profiles.

In summary, we demonstrated that \themethod{} achieves best-in-class performance in addressing several biological discovery tasks by deriving knowledge from diverse pooled perturbation experiments - opening up a new and promising direction of research on integrative perturbation models for biological discovery that harness transfer learning from heterogeneous experimental data. \themethod{} could be of high utility to conduct experimental studies that are predictable in computers from existing data in lieu of expensive and time-consuming laboratory experiments, and could complement real-world laboratory experiments by focusing experimental data generation on data points that are most uncertain given existing knowledge.
 
\section*{Data and code availability}
\themethod{} instances are trained on data from publicly available perturbation screens from \citet{norman2019exploring}, \citet{replogle2022mapping} and \citet{subramanian2017next} (available at \url{https://clue.io/}). Source code, pretrained \themethod{}s and embeddings for downstream tasks will be made available upon publication. The Optum\textsuperscript{\textcopyright} de-identified Electronic Health Record database used to validate in-silico findings in real-world data is available for accredited researchers from Optum, Inc. but third-party restrictions apply to the availability of these data. The data were used under license for this study with restrictions that do not allow for the data to be redistributed or made publicly available. Data access to the Optum\textsuperscript{\textcopyright} de-identified Electronic Health Record database may require a data sharing agreement and may incur data access fees.

\section*{Acknowledgements}
{\DJ}M, MC, AG, LS, AM, MB and PS are employees and shareholders of GSK plc. TH is a former employee of GSK plc. We thank Xue Long Zhao, Caroline Weis, and Stefan Bauer for their valuable feedback.

\nocite{florian_charlier_2022_7213391} %

\bibliographystyle{unsrtnat}
\bibliography{references}

\begin{thebibliography}{85}
\providecommand{\natexlab}[1]{#1}
\providecommand{\url}[1]{\texttt{#1}}
\expandafter\ifx\csname urlstyle\endcsname\relax
  \providecommand{\doi}[1]{doi: #1}\else
  \providecommand{\doi}{doi: \begingroup \urlstyle{rm}\Url}\fi

\bibitem[Meinshausen et~al.(2016)Meinshausen, Hauser, Mooij, Peters, Versteeg,
  and B{\"u}hlmann]{meinshausen2016methods}
Nicolai Meinshausen, Alain Hauser, Joris~M Mooij, Jonas Peters, Philip
  Versteeg, and Peter B{\"u}hlmann.
\newblock Methods for causal inference from gene perturbation experiments and
  validation.
\newblock \emph{Proceedings of the National Academy of Sciences}, 113\penalty0
  (27):\penalty0 7361--7368, 2016.

\bibitem[Rubin et~al.(2019)Rubin, Parker, Satpathy, Qi, Wu, Ong, Mumbach, Ji,
  Kim, Cho, et~al.]{rubin2019coupled}
Adam~J Rubin, Kevin~R Parker, Ansuman~T Satpathy, Yanyan Qi, Beijing Wu,
  Alvin~J Ong, Maxwell~R Mumbach, Andrew~L Ji, Daniel~S Kim, Seung~Woo Cho,
  et~al.
\newblock {Coupled single-cell CRISPR screening and epigenomic profiling
  reveals causal gene regulatory networks}.
\newblock \emph{Cell}, 176\penalty0 (1):\penalty0 361--376, 2019.

\bibitem[Tejada-Lapuerta et~al.(2023)Tejada-Lapuerta, Bertin, Bauer, Aliee,
  Bengio, and Theis]{tejada2023causal}
Alejandro Tejada-Lapuerta, Paul Bertin, Stefan Bauer, Hananeh Aliee, Yoshua
  Bengio, and Fabian~J Theis.
\newblock Causal machine learning for single-cell genomics.
\newblock \emph{arXiv preprint arXiv:2310.14935}, 2023.

\bibitem[Biermann et~al.(2017)Biermann, Kanie, and Kim]{biermann2017global}
Frank Biermann, Norichika Kanie, and Rakhyun~E Kim.
\newblock Global governance by goal-setting: the novel approach of the un
  sustainable development goals.
\newblock \emph{Current Opinion in Environmental Sustainability}, 26:\penalty0
  26--31, 2017.

\bibitem[Shalem et~al.(2015)Shalem, Sanjana, and Zhang]{shalem2015high}
Ophir Shalem, Neville~E Sanjana, and Feng Zhang.
\newblock {High-throughput functional genomics using CRISPR--Cas9}.
\newblock \emph{Nature Reviews Genetics}, 16\penalty0 (5):\penalty0 299--311,
  2015.

\bibitem[Rauscher et~al.(2016)Rauscher, Heigwer, Breinig, Winter, and
  Boutros]{rauscher2016genomecrispr}
Benedikt Rauscher, Florian Heigwer, Marco Breinig, Jan Winter, and Michael
  Boutros.
\newblock {GenomeCRISPR-a database for high-throughput CRISPR/Cas9 screens}.
\newblock \emph{Nucleic acids research}, page gkw997, 2016.

\bibitem[Subramanian et~al.(2017)Subramanian, Narayan, Corsello, Peck, Natoli,
  Lu, Gould, Davis, Tubelli, Asiedu, et~al.]{subramanian2017next}
Aravind Subramanian, Rajiv Narayan, Steven~M Corsello, David~D Peck, Ted~E
  Natoli, Xiaodong Lu, Joshua Gould, John~F Davis, Andrew~A Tubelli, Jacob~K
  Asiedu, et~al.
\newblock {A next generation connectivity map: L1000 platform and the first
  1,000,000 profiles}.
\newblock \emph{Cell}, 171\penalty0 (6):\penalty0 1437--1452, 2017.

\bibitem[Oughtred et~al.(2019)Oughtred, Stark, Breitkreutz, Rust, Boucher,
  Chang, Kolas, O’Donnell, Leung, McAdam, et~al.]{oughtred2019biogrid}
Rose Oughtred, Chris Stark, Bobby-Joe Breitkreutz, Jennifer Rust, Lorrie
  Boucher, Christie Chang, Nadine Kolas, Lara O’Donnell, Genie Leung,
  Rochelle McAdam, et~al.
\newblock The biogrid interaction database: 2019 update.
\newblock \emph{Nucleic acids research}, 47\penalty0 (D1):\penalty0 D529--D541,
  2019.

\bibitem[Replogle et~al.(2022)Replogle, Saunders, Pogson, Hussmann, Lenail,
  Guna, Mascibroda, Wagner, Adelman, Lithwick-Yanai,
  et~al.]{replogle2022mapping}
Joseph~M Replogle, Reuben~A Saunders, Angela~N Pogson, Jeffrey~A Hussmann,
  Alexander Lenail, Alina Guna, Lauren Mascibroda, Eric~J Wagner, Karen
  Adelman, Gila Lithwick-Yanai, et~al.
\newblock Mapping information-rich genotype-phenotype landscapes with
  genome-scale perturb-seq.
\newblock \emph{Cell}, 185\penalty0 (14):\penalty0 2559--2575, 2022.

\bibitem[Kamimoto et~al.(2023)Kamimoto, Stringa, Hoffmann, Jindal,
  Solnica-Krezel, and Morris]{kamimoto2023dissecting}
Kenji Kamimoto, Blerta Stringa, Christy~M Hoffmann, Kunal Jindal, Lilianna
  Solnica-Krezel, and Samantha~A Morris.
\newblock Dissecting cell identity via network inference and in silico gene
  perturbation.
\newblock \emph{Nature}, 614\penalty0 (7949):\penalty0 742--751, 2023.

\bibitem[Roohani et~al.(2023)Roohani, Huang, and
  Leskovec]{roohani2023predicting}
Yusuf Roohani, Kexin Huang, and Jure Leskovec.
\newblock {Predicting transcriptional outcomes of novel multigene perturbations
  with GEARS}.
\newblock \emph{Nature Biotechnology}, pages 1--9, 2023.

\bibitem[Yuan et~al.(2021)Yuan, Shen, Luna, Korkut, Marks, Ingraham, and
  Sander]{yuan2021cellbox}
Bo~Yuan, Ciyue Shen, Augustin Luna, Anil Korkut, Debora~S Marks, John Ingraham,
  and Chris Sander.
\newblock Cellbox: interpretable machine learning for perturbation biology with
  application to the design of cancer combination therapy.
\newblock \emph{Cell systems}, 12\penalty0 (2):\penalty0 128--140, 2021.

\bibitem[Lotfollahi et~al.(2019)Lotfollahi, Wolf, and
  Theis]{lotfollahi2019scgen}
Mohammad Lotfollahi, F~Alexander Wolf, and Fabian~J Theis.
\newblock {scGen predicts single-cell perturbation responses}.
\newblock \emph{Nature methods}, 16\penalty0 (8):\penalty0 715--721, 2019.

\bibitem[Hetzel et~al.(2022)Hetzel, Boehm, Kilbertus, G{\"u}nnemann, Theis,
  et~al.]{hetzel2022predicting}
Leon Hetzel, Simon Boehm, Niki Kilbertus, Stephan G{\"u}nnemann, Fabian Theis,
  et~al.
\newblock Predicting cellular responses to novel drug perturbations at a
  single-cell resolution.
\newblock \emph{Advances in Neural Information Processing Systems},
  35:\penalty0 26711--26722, 2022.

\bibitem[Lotfollahi et~al.(2023)Lotfollahi, Klimovskaia~Susmelj, De~Donno,
  Hetzel, Ji, Ibarra, Srivatsan, Naghipourfar, Daza, Martin,
  et~al.]{lotfollahi2023predicting}
Mohammad Lotfollahi, Anna Klimovskaia~Susmelj, Carlo De~Donno, Leon Hetzel,
  Yuge Ji, Ignacio~L Ibarra, Sanjay~R Srivatsan, Mohsen Naghipourfar, Riza~M
  Daza, Beth Martin, et~al.
\newblock Predicting cellular responses to complex perturbations in
  high-throughput screens.
\newblock \emph{Molecular Systems Biology}, page e11517, 2023.

\bibitem[Wu et~al.(2022{\natexlab{a}})Wu, Barton, Wang, Ioannidis, De~Donno,
  Price, Voloch, and Karypis]{wu2022predicting}
Yulun Wu, Robert~A Barton, Zichen Wang, Vassilis~N Ioannidis, Carlo De~Donno,
  Layne~C Price, Luis~F Voloch, and George Karypis.
\newblock Predicting cellular responses with variational causal inference and
  refined relational information.
\newblock \emph{arXiv preprint arXiv:2210.00116}, 2022{\natexlab{a}}.

\bibitem[Bunne et~al.(2023)Bunne, Stark, Gut, Del~Castillo, Levesque, Lehmann,
  Pelkmans, Krause, and R{\"a}tsch]{bunne2023learning}
Charlotte Bunne, Stefan~G Stark, Gabriele Gut, Jacobo~Sarabia Del~Castillo,
  Mitch Levesque, Kjong-Van Lehmann, Lucas Pelkmans, Andreas Krause, and Gunnar
  R{\"a}tsch.
\newblock Learning single-cell perturbation responses using neural optimal
  transport.
\newblock \emph{Nature methods}, 20\penalty0 (11):\penalty0 1759--1768, 2023.

\bibitem[Chevalley et~al.(2022)Chevalley, Roohani, Mehrjou, Leskovec, and
  Schwab]{chevalley2022causalbench}
Mathieu Chevalley, Yusuf Roohani, Arash Mehrjou, Jure Leskovec, and Patrick
  Schwab.
\newblock {CausalBench: A Large-scale Benchmark for Network Inference from
  Single-cell Perturbation Data}.
\newblock \emph{arXiv preprint arXiv:2210.17283}, 2022.

\bibitem[Lopez et~al.(2022)Lopez, Tagasovska, Ra, Cho, Pritchard, and
  Regev]{lopez2022learning}
Romain Lopez, Nata{\v{s}}a Tagasovska, Stephen Ra, Kyunghyn Cho, Jonathan~K
  Pritchard, and Aviv Regev.
\newblock Learning causal representations of single cells via sparse mechanism
  shift modeling.
\newblock \emph{arXiv preprint arXiv:2211.03553}, 2022.

\bibitem[Rosen et~al.(2023)Rosen, Roohani, Agarwal, Samotor{\v{c}}an,
  Consortium, Quake, and Leskovec]{rosen2023universal}
Yanay Rosen, Yusuf Roohani, Ayush Agarwal, Leon Samotor{\v{c}}an,
  Tabula~Sapiens Consortium, Stephen~R Quake, and Jure Leskovec.
\newblock Universal cell embeddings: A foundation model for cell biology.
\newblock \emph{bioRxiv}, pages 2023--11, 2023.

\bibitem[Schmauch et~al.(2020)Schmauch, Romagnoni, Pronier, Saillard,
  Maill{\'e}, Calderaro, Kamoun, Sefta, Toldo, Zaslavskiy,
  et~al.]{schmauch2020deep}
Beno{\^\i}t Schmauch, Alberto Romagnoni, Elodie Pronier, Charlie Saillard,
  Pascale Maill{\'e}, Julien Calderaro, Aur{\'e}lie Kamoun, Meriem Sefta,
  Sylvain Toldo, Mikhail Zaslavskiy, et~al.
\newblock A deep learning model to predict rna-seq expression of tumours from
  whole slide images.
\newblock \emph{Nature communications}, 11\penalty0 (1):\penalty0 3877, 2020.

\bibitem[Arslan et~al.(2022)Arslan, Mehrotra, Schmidt, Geraldes, Singhal,
  Hense, Li, Bass, and Raharja-Liu]{arslan2022large}
Salim Arslan, Debapriya Mehrotra, Julian Schmidt, Andre Geraldes, Shikha
  Singhal, Julius Hense, Xiusi Li, Cher Bass, and Pandu Raharja-Liu.
\newblock Large-scale systematic feasibility study on the pan-cancer
  predictability of multi-omic biomarkers from whole slide images with deep
  learning.
\newblock \emph{bioRxiv}, pages 2022--01, 2022.

\bibitem[Mehrizi et~al.(2023)Mehrizi, Mehrjou, Alegro, Zhao, Carbone, Fishwick,
  Vappiani, Bi, Sanford, Keles, Bantscheff, Nguyen, and
  Schwab]{mehrizi2023multiomics}
Rahil Mehrizi, Arash Mehrjou, Maryana Alegro, Yi~Zhao, Benedetta Carbone, Carl
  Fishwick, Johanna Vappiani, Jing Bi, Siobhan Sanford, Hakan Keles, Marcus
  Bantscheff, Cuong Nguyen, and Patrick Schwab.
\newblock Multi-omics prediction from high-content cellular imaging with deep
  learning.
\newblock \emph{arXiv preprint arXiv:306.09391}, 2023.

\bibitem[Mehrjou et~al.(2022)Mehrjou, Soleymani, Jesson, Notin, Gal, Bauer, and
  Schwab]{mehrjou2021genedisco}
Arash Mehrjou, Ashkan Soleymani, Andrew Jesson, Pascal Notin, Yarin Gal, Stefan
  Bauer, and Patrick Schwab.
\newblock {GeneDisco: A Benchmark for Experimental Design in Drug Discovery}.
\newblock In \emph{{International Conference on Learning Representations}},
  2022.

\bibitem[Lyle et~al.(2023)Lyle, Mehrjou, Notin, Jesson, Bauer, Gal, and
  Schwab]{lyle2023discobax}
Clare Lyle, Arash Mehrjou, Pascal Notin, Andrew Jesson, Stefan Bauer, Yarin
  Gal, and Patrick Schwab.
\newblock {D}isco{BAX} discovery of optimal intervention sets in genomic
  experiment design.
\newblock In \emph{Proceedings of the 40th International Conference on Machine
  Learning}, volume 202 of \emph{Proceedings of Machine Learning Research},
  pages 23170--23189. PMLR, 23--29 Jul 2023.
\newblock URL \url{https://proceedings.mlr.press/v202/lyle23a.html}.

\bibitem[Theodoris et~al.(2023)Theodoris, Xiao, Chopra, Chaffin, Al~Sayed,
  Hill, Mantineo, Brydon, Zeng, Liu, et~al.]{theodoris2023transfer}
Christina~V Theodoris, Ling Xiao, Anant Chopra, Mark~D Chaffin, Zeina~R
  Al~Sayed, Matthew~C Hill, Helene Mantineo, Elizabeth~M Brydon, Zexian Zeng,
  X~Shirley Liu, et~al.
\newblock Transfer learning enables predictions in network biology.
\newblock \emph{Nature}, pages 1--9, 2023.

\bibitem[Cui et~al.(2023)Cui, Wang, Maan, Pang, Luo, and Wang]{cui2023scgpt}
Haotian Cui, Chloe Wang, Hassaan Maan, Kuan Pang, Fengning Luo, and Bo~Wang.
\newblock scgpt: Towards building a foundation model for single-cell
  multi-omics using generative ai.
\newblock \emph{bioRxiv}, pages 2023--04, 2023.

\bibitem[Hao et~al.(2023)Hao, Gong, Zeng, Liu, Guo, Cheng, Wang, Ma, Song, and
  Zhang]{hao2023large}
Minsheng Hao, Jing Gong, Xin Zeng, Chiming Liu, Yucheng Guo, Xingyi Cheng,
  Taifeng Wang, Jianzhu Ma, Le~Song, and Xuegong Zhang.
\newblock Large scale foundation model on single-cell transcriptomics.
\newblock \emph{bioRxiv}, pages 2023--05, 2023.

\bibitem[Chen and Zou(2023)]{chen2023genept}
Yiqun~T Chen and James Zou.
\newblock {GenePT: A simple but hard-to-beat foundation model for genes and
  cells built from ChatGPT}.
\newblock \emph{bioRxiv}, pages 2023--10, 2023.

\bibitem[Norman et~al.(2019)Norman, Horlbeck, Replogle, Ge, Xu, Jost, Gilbert,
  and Weissman]{norman2019exploring}
Thomas~M Norman, Max~A Horlbeck, Joseph~M Replogle, Alex~Y Ge, Albert Xu, Marco
  Jost, Luke~A Gilbert, and Jonathan~S Weissman.
\newblock Exploring genetic interaction manifolds constructed from rich
  single-cell phenotypes.
\newblock \emph{Science}, 365\penalty0 (6455):\penalty0 786--793, 2019.

\bibitem[Prokhorenkova et~al.(2018)Prokhorenkova, Gusev, Vorobev, Dorogush, and
  Gulin]{prokhorenkova2018catboost}
Liudmila Prokhorenkova, Gleb Gusev, Aleksandr Vorobev, Anna~Veronika Dorogush,
  and Andrey Gulin.
\newblock Catboost: unbiased boosting with categorical features.
\newblock \emph{Advances in neural information processing systems}, 31, 2018.

\bibitem[Szklarczyk et~al.(2019)Szklarczyk, Gable, Lyon, Junge, Wyder,
  Huerta-Cepas, Simonovic, Doncheva, Morris, Bork,
  et~al.]{szklarczyk2019string}
Damian Szklarczyk, Annika~L Gable, David Lyon, Alexander Junge, Stefan Wyder,
  Jaime Huerta-Cepas, Milan Simonovic, Nadezhda~T Doncheva, John~H Morris, Peer
  Bork, et~al.
\newblock String v11: protein--protein association networks with increased
  coverage, supporting functional discovery in genome-wide experimental
  datasets.
\newblock \emph{Nucleic acids research}, 47\penalty0 (D1):\penalty0 D607--D613,
  2019.

\bibitem[Fabregat et~al.(2018)Fabregat, Jupe, Matthews, Sidiropoulos,
  Gillespie, Garapati, Haw, Jassal, Korninger, May,
  et~al.]{fabregat2018reactome}
Antonio Fabregat, Steven Jupe, Lisa Matthews, Konstantinos Sidiropoulos, Marc
  Gillespie, Phani Garapati, Robin Haw, Bijay Jassal, Florian Korninger, Bruce
  May, et~al.
\newblock The reactome pathway knowledgebase.
\newblock \emph{Nucleic acids research}, 46\penalty0 (D1):\penalty0 D649--D655,
  2018.

\bibitem[Du et~al.(2019)Du, Jia, Dai, Tao, Zhao, and Zhi]{du2019gene2vec}
Jingcheng Du, Peilin Jia, Yulin Dai, Cui Tao, Zhongming Zhao, and Degui Zhi.
\newblock Gene2vec: distributed representation of genes based on co-expression.
\newblock \emph{BMC genomics}, 20:\penalty0 7--15, 2019.

\bibitem[Horlbeck et~al.(2018)Horlbeck, Xu, Wang, Bennett, Park, Bogdanoff,
  Adamson, Chow, Kampmann, Peterson, et~al.]{horlbeck2018mapping}
Max~A Horlbeck, Albert Xu, Min Wang, Neal~K Bennett, Chong~Y Park, Derek
  Bogdanoff, Britt Adamson, Eric~D Chow, Martin Kampmann, Tim~R Peterson,
  et~al.
\newblock Mapping the genetic landscape of human cells.
\newblock \emph{Cell}, 174\penalty0 (4):\penalty0 953--967, 2018.

\bibitem[Van~der Maaten and Hinton(2008)]{van2008visualizing}
Laurens Van~der Maaten and Geoffrey Hinton.
\newblock {Visualizing data using t-SNE}.
\newblock \emph{Journal of machine learning research}, 9\penalty0 (11), 2008.

\bibitem[Tribouilloy et~al.(2011)Tribouilloy, Jeu, Mar{\'e}chaux, Jobic,
  Rusinaru, and Andrejak]{tribouilloy2011benfluorex}
Christophe Tribouilloy, Antoine Jeu, Sylvestre Mar{\'e}chaux, Yannick Jobic,
  Dan Rusinaru, and Michel Andrejak.
\newblock Benfluorex and valvular heart disease.
\newblock \emph{Presse Medicale (Paris, France: 1983)}, 40\penalty0
  (11):\penalty0 1008--1016, 2011.

\bibitem[Jiang et~al.(2023)Jiang, Hu, McIntosh, and
  Shah]{jiang2023investigating}
Jiayue-Clara Jiang, Chenwen Hu, Andrew~M McIntosh, and Sonia Shah.
\newblock Investigating the potential anti-depressive mechanisms of statins: a
  transcriptomic and mendelian randomization analysis.
\newblock \emph{Translational Psychiatry}, 13\penalty0 (1):\penalty0 110, 2023.

\bibitem[Blake and Ridker(2000)]{blake2000statins}
Gavin~J Blake and Paul~M Ridker.
\newblock Are statins anti-inflammatory?
\newblock \emph{Trials}, 1:\penalty0 1--5, 2000.

\bibitem[McGown et~al.(2010)McGown, Brown, Hellewell, Reilly, and
  Brookes]{mcgown2010beneficial}
CC~McGown, NJ~Brown, PG~Hellewell, CS~Reilly, and ZLS Brookes.
\newblock {Beneficial microvascular and anti-inflammatory effects of
  pravastatin during sepsis involve nitric oxide synthase III}.
\newblock \emph{British journal of anaesthesia}, 104\penalty0 (2):\penalty0
  183--190, 2010.

\bibitem[Sommeijer et~al.(2004)Sommeijer, MacGillavry, Meijers, Van~Zanten,
  Reitsma, and Cate]{sommeijer2004anti}
Dirkje~W Sommeijer, Melvin~R MacGillavry, Joost~CM Meijers, Anton~P Van~Zanten,
  Pieter~H Reitsma, and Hugo~Ten Cate.
\newblock Anti-inflammatory and anticoagulant effects of pravastatin in
  patients with type 2 diabetes.
\newblock \emph{Diabetes care}, 27\penalty0 (2):\penalty0 468--473, 2004.

\bibitem[Jiang et~al.(2021)Jiang, Prabhakar, Van~der Voorn, Ghatpande, Celona,
  Venkataramanan, Calviello, Lin, Wang, Black, et~al.]{jiang2021control}
Xuan Jiang, Amit Prabhakar, Stephanie~M Van~der Voorn, Prajakta Ghatpande,
  Barbara Celona, Srivats Venkataramanan, Lorenzo Calviello, Chuwen Lin,
  Wanpeng Wang, Brian~L Black, et~al.
\newblock Control of ribosomal protein synthesis by the microprocessor complex.
\newblock \emph{Science signaling}, 14\penalty0 (671):\penalty0 eabd2639, 2021.

\bibitem[Giurgiu et~al.(2019)Giurgiu, Reinhard, Brauner, Dunger-Kaltenbach,
  Fobo, Frishman, Montrone, and Ruepp]{giurgiu2019corum}
Madalina Giurgiu, Julian Reinhard, Barbara Brauner, Irmtraud Dunger-Kaltenbach,
  Gisela Fobo, Goar Frishman, Corinna Montrone, and Andreas Ruepp.
\newblock Corum: the comprehensive resource of mammalian protein
  complexes—2019.
\newblock \emph{Nucleic acids research}, 47\penalty0 (D1):\penalty0 D559--D563,
  2019.

\bibitem[Hopp et~al.(2012)Hopp, Ward, Hommerding, Nasr, Tuan, Gainullin,
  Rossetti, Torres, Harris, et~al.]{hopp2012functional}
Katharina Hopp, Christopher~J Ward, Cynthia~J Hommerding, Samih~H Nasr,
  Han-Fang Tuan, Vladimir~G Gainullin, Sandro Rossetti, Vicente~E Torres,
  Peter~C Harris, et~al.
\newblock Functional polycystin-1 dosage governs autosomal dominant polycystic
  kidney disease severity.
\newblock \emph{The Journal of clinical investigation}, 122\penalty0
  (11):\penalty0 4257--4273, 2012.

\bibitem[Rossetti et~al.(2009)Rossetti, Kubly, Consugar, Hopp, Roy, Horsley,
  Chauveau, Rees, Barratt, Van't~Hoff, et~al.]{rossetti2009incompletely}
Sandro Rossetti, Vickie~J Kubly, Mark~B Consugar, Katharina Hopp, Sushmita Roy,
  Sharon~W Horsley, Dominique Chauveau, Lesley Rees, T~Martin Barratt,
  William~G Van't~Hoff, et~al.
\newblock {Incompletely penetrant PKD1 alleles suggest a role for gene dosage
  in cyst initiation in polycystic kidney disease}.
\newblock \emph{Kidney international}, 75\penalty0 (8):\penalty0 848--855,
  2009.

\bibitem[Gainullin et~al.(2015)Gainullin, Hopp, Ward, Hommerding, Harris,
  et~al.]{gainullin2015polycystin}
Vladimir~G Gainullin, Katharina Hopp, Christopher~J Ward, Cynthia~J Hommerding,
  Peter~C Harris, et~al.
\newblock Polycystin-1 maturation requires polycystin-2 in a dose-dependent
  manner.
\newblock \emph{The Journal of clinical investigation}, 125\penalty0
  (2):\penalty0 607--620, 2015.

\bibitem[Torres et~al.(2012)Torres, Chapman, Devuyst, Gansevoort, Grantham,
  Higashihara, Perrone, Krasa, Ouyang, and Czerwiec]{torres2012tolvaptan}
Vicente~E Torres, Arlene~B Chapman, Olivier Devuyst, Ron~T Gansevoort, Jared~J
  Grantham, Eiji Higashihara, Ronald~D Perrone, Holly~B Krasa, John Ouyang, and
  Frank~S Czerwiec.
\newblock Tolvaptan in patients with autosomal dominant polycystic kidney
  disease.
\newblock \emph{New England Journal of Medicine}, 367\penalty0 (25):\penalty0
  2407--2418, 2012.

\bibitem[Wang et~al.(2008)Wang, Wu, Ward, Harris, and
  Torres]{wang2008vasopressin}
Xiaofang Wang, Yanhong Wu, Christopher~J Ward, Peter~C Harris, and Vicente~E
  Torres.
\newblock Vasopressin directly regulates cyst growth in polycystic kidney
  disease.
\newblock \emph{Journal of the American Society of Nephrology: JASN},
  19\penalty0 (1):\penalty0 102, 2008.

\bibitem[Reeders et~al.(1985)Reeders, Breuning, Davies, Nicholls, Jarman,
  Higgs, Pearson, and Weatherall]{reeders1985highly}
ST~Reeders, MH~Breuning, KE~Davies, RD~Nicholls, AP~Jarman, DR~Higgs,
  PL~Pearson, and DJ~Weatherall.
\newblock {A highly polymorphic DNA marker linked to adult polycystic kidney
  disease on chromosome 16}.
\newblock \emph{Nature}, 317\penalty0 (6037):\penalty0 542--544, 1985.

\bibitem[Lanktree et~al.(2021)Lanktree, Haghighi, di~Bari, Song, and
  Pei]{lanktree2021insights}
Matthew~B Lanktree, Amirreza Haghighi, Ighli di~Bari, Xuewen Song, and York
  Pei.
\newblock Insights into autosomal dominant polycystic kidney disease from
  genetic studies.
\newblock \emph{Clinical Journal of the American Society of Nephrology: CJASN},
  16\penalty0 (5):\penalty0 790, 2021.

\bibitem[Radhakrishnan et~al.(2022)Radhakrishnan, Duriseti, and
  Chebib]{radhakrishnan2022management}
Yeshwanter Radhakrishnan, Parikshit Duriseti, and Fouad~T Chebib.
\newblock Management of autosomal dominant polycystic kidney disease in the era
  of disease-modifying treatment options.
\newblock \emph{Kidney Research and Clinical Practice}, 41\penalty0
  (4):\penalty0 422, 2022.

\bibitem[Duan et~al.(2016)Duan, Reid, Clark, Wang, Fernandez, Rouillard,
  Readhead, Tritsch, Hodos, Hafner, et~al.]{duan2016l1000cds2}
Qiaonan Duan, St~Patrick Reid, Neil~R Clark, Zichen Wang, Nicolas~F Fernandez,
  Andrew~D Rouillard, Ben Readhead, Sarah~R Tritsch, Rachel Hodos, Marc Hafner,
  et~al.
\newblock {L1000CDS2: LINCS L1000 characteristic direction signatures search
  engine}.
\newblock \emph{NPJ systems biology and applications}, 2\penalty0 (1):\penalty0
  1--12, 2016.

\bibitem[Gbelcov{\'a} et~al.(2017)Gbelcov{\'a}, Rimpelov{\'a}, Ruml,
  Fenclov{\'a}, Kosek, Haj{\v{s}}lov{\'a}, Strnad, Kol{\'a}{\v{r}}, and
  V{\'\i}tek]{gbelcova2017variability}
Helena Gbelcov{\'a}, Silvie Rimpelov{\'a}, Tom{\'a}{\v{s}} Ruml, Marie
  Fenclov{\'a}, V{\'\i}tek Kosek, Jana Haj{\v{s}}lov{\'a}, Hynek Strnad, Michal
  Kol{\'a}{\v{r}}, and Libor V{\'\i}tek.
\newblock Variability in statin-induced changes in gene expression profiles of
  pancreatic cancer.
\newblock \emph{Scientific Reports}, 7\penalty0 (1):\penalty0 44219, 2017.

\bibitem[Huang et~al.(2023)Huang, Wu, Wu, Li, Tan, Shen, Xiong, Feng, Gao, Li,
  et~al.]{huang2023long}
Tong-sheng Huang, Teng Wu, Yan-di Wu, Xing-hui Li, Jing Tan, Cong-hui Shen,
  Shi-jie Xiong, Zi-qi Feng, Sai-fei Gao, Hui Li, et~al.
\newblock Long-term statins administration exacerbates diabetic nephropathy via
  ectopic fat deposition in diabetic mice.
\newblock \emph{Nature Communications}, 14\penalty0 (1):\penalty0 390, 2023.

\bibitem[Cadnapaphornchai et~al.(2014)Cadnapaphornchai, George, McFann, Wang,
  Gitomer, Strain, and Schrier]{cadnapaphornchai2014effect}
Melissa~A Cadnapaphornchai, Diana~M George, Kim McFann, Wei Wang, Berenice
  Gitomer, John~D Strain, and Robert~W Schrier.
\newblock Effect of pravastatin on total kidney volume, left ventricular mass
  index, and microalbuminuria in pediatric autosomal dominant polycystic kidney
  disease.
\newblock \emph{Clinical journal of the American Society of Nephrology: CJASN},
  9\penalty0 (5):\penalty0 889, 2014.

\bibitem[Leuenroth et~al.(2007)Leuenroth, Okuhara, Shotwell, Markowitz, Yu,
  Somlo, and Crews]{leuenroth2007triptolide}
Stephanie~J Leuenroth, Dayne Okuhara, Joseph~D Shotwell, Glen~S Markowitz,
  Zhiheng Yu, Stefan Somlo, and Craig~M Crews.
\newblock Triptolide is a traditional chinese medicine-derived inhibitor of
  polycystic kidney disease.
\newblock \emph{Proceedings of the National Academy of Sciences}, 104\penalty0
  (11):\penalty0 4389--4394, 2007.

\bibitem[Leuenroth et~al.(2008)Leuenroth, Bencivenga, Igarashi, Somlo, and
  Crews]{leuenroth2008triptolide}
Stephanie~J Leuenroth, Natasha Bencivenga, Peter Igarashi, Stefan Somlo, and
  Craig~M Crews.
\newblock {Triptolide reduces cystogenesis in a model of ADPKD}.
\newblock \emph{Journal of the American Society of Nephrology: JASN},
  19\penalty0 (9):\penalty0 1659, 2008.

\bibitem[Deng and Guan(2023)]{deng2023supervised}
Kaiwen Deng and Yuanfang Guan.
\newblock {A Supervised LightGBM-Based Approach to the GSK.ai CausalBench
  Challenge (ICLR 2023)}.
\newblock 2023.
\newblock URL \url{https://openreview.net/forum?id=nB9zUwS2gpI}.

\bibitem[Kaplan et~al.(2020)Kaplan, McCandlish, Henighan, Brown, Chess, Child,
  Gray, Radford, Wu, and Amodei]{kaplan2020scaling}
Jared Kaplan, Sam McCandlish, Tom Henighan, Tom~B Brown, Benjamin Chess, Rewon
  Child, Scott Gray, Alec Radford, Jeffrey Wu, and Dario Amodei.
\newblock Scaling laws for neural language models.
\newblock \emph{arXiv preprint arXiv:2001.08361}, 2020.

\bibitem[Hoffmann et~al.(2022)Hoffmann, Borgeaud, Mensch, Buchatskaya, Cai,
  Rutherford, Casas, Hendricks, Welbl, Clark, et~al.]{hoffmann2022training}
Jordan Hoffmann, Sebastian Borgeaud, Arthur Mensch, Elena Buchatskaya, Trevor
  Cai, Eliza Rutherford, Diego de~Las Casas, Lisa~Anne Hendricks, Johannes
  Welbl, Aidan Clark, et~al.
\newblock Training compute-optimal large language models.
\newblock \emph{arXiv preprint arXiv:2203.15556}, 2022.

\bibitem[Leek et~al.(2010)Leek, Scharpf, Bravo, Simcha, Langmead, Johnson,
  Geman, Baggerly, and Irizarry]{leek2010tackling}
Jeffrey~T Leek, Robert~B Scharpf, H{\'e}ctor~Corrada Bravo, David Simcha,
  Benjamin Langmead, W~Evan Johnson, Donald Geman, Keith Baggerly, and Rafael~A
  Irizarry.
\newblock Tackling the widespread and critical impact of batch effects in
  high-throughput data.
\newblock \emph{Nature Reviews Genetics}, 11\penalty0 (10):\penalty0 733--739,
  2010.

\bibitem[Errington et~al.(2021)Errington, Mathur, Soderberg, Denis, Perfito,
  Iorns, and Nosek]{errington2021investigating}
Timothy~M Errington, Maya Mathur, Courtney~K Soderberg, Alexandria Denis,
  Nicole Perfito, Elizabeth Iorns, and Brian~A Nosek.
\newblock Investigating the replicability of preclinical cancer biology.
\newblock \emph{Elife}, 10:\penalty0 e71601, 2021.

\bibitem[Schwab et~al.(2020)Schwab, Linhardt, Bauer, Buhmann, and
  Karlen]{schwab2020doseresponse}
Patrick Schwab, Lorenz Linhardt, Stefan Bauer, Joachim~M Buhmann, and Walter
  Karlen.
\newblock {Learning Counterfactual Representations for Estimating Individual
  Dose-Response Curves}.
\newblock In \emph{{AAAI Conference on Artificial Intelligence}}, 2020.

\bibitem[Charlier et~al.(2022)Charlier, Weber, Izak, Harkin, Magnus, Lalli,
  Fresnais, Chan, Markov, Amsalem, Proost, Krasoulis, getzze, and
  Repplinger]{florian_charlier_2022_7213391}
Florian Charlier, Marc Weber, Dariusz Izak, Emerson Harkin, Marcin Magnus,
  Joseph Lalli, Louison Fresnais, Matt Chan, Nikolay Markov, Oren Amsalem,
  Sebastian Proost, Agamemnon Krasoulis, getzze, and Stefan Repplinger.
\newblock Statannotations, October 2022.
\newblock URL \url{https://doi.org/10.5281/zenodo.7213391}.

\bibitem[Wu et~al.(2022{\natexlab{b}})Wu, Price, Wang, Ioannidis, and
  Karypis]{wu2022variational}
Yulun Wu, Layne~C Price, Zichen Wang, Vassilis~N Ioannidis, and George Karypis.
\newblock Variational causal inference.
\newblock \emph{arXiv preprint arXiv:2209.05935}, 2022{\natexlab{b}}.

\bibitem[Ye et~al.(2018)Ye, Ho, Neri, Yang, Kulkarni, Randhawa, Henault,
  Mostacci, Farmer, Renner, et~al.]{ye2018drug}
Chaoyang Ye, Daniel~J Ho, Marilisa Neri, Chian Yang, Tripti Kulkarni, Ranjit
  Randhawa, Martin Henault, Nadezda Mostacci, Pierre Farmer, Steffen Renner,
  et~al.
\newblock Drug-seq for miniaturized high-throughput transcriptome profiling in
  drug discovery.
\newblock \emph{Nature communications}, 9\penalty0 (1):\penalty0 4307, 2018.

\bibitem[Tsherniak et~al.(2017)Tsherniak, Vazquez, Montgomery, Weir, Kryukov,
  Cowley, Gill, Harrington, Pantel, Krill-Burger,
  et~al.]{tsherniak2017defining}
Aviad Tsherniak, Francisca Vazquez, Phil~G Montgomery, Barbara~A Weir, Gregory
  Kryukov, Glenn~S Cowley, Stanley Gill, William~F Harrington, Sasha Pantel,
  John~M Krill-Burger, et~al.
\newblock Defining a cancer dependency map.
\newblock \emph{Cell}, 170\penalty0 (3):\penalty0 564--576, 2017.

\bibitem[Kingma and Ba(2014)]{kingma2014adam}
Diederik~P Kingma and Jimmy Ba.
\newblock Adam: A method for stochastic optimization.
\newblock \emph{arXiv preprint arXiv:1412.6980}, 2014.

\bibitem[Chevalley et~al.(2023)Chevalley, Sackett-Sanders, Roohani, Notin,
  Bakulin, Brzezinski, Deng, Guan, Hong, Ibrahim, Kotlowski, Kowiel, Misiakos,
  Nazaret, Püschel, Wendler, Mehrjou, and Schwab]{chevalley2023causalbench}
Mathieu Chevalley, Jacob Sackett-Sanders, Yusuf Roohani, Pascal Notin, Artemy
  Bakulin, Dariusz Brzezinski, Kaiwen Deng, Yuanfang Guan, Justin Hong, Michael
  Ibrahim, Wojciech Kotlowski, Marcin Kowiel, Panagiotis Misiakos, Achille
  Nazaret, Markus Püschel, Chris Wendler, Arash Mehrjou, and Patrick Schwab.
\newblock {The CausalBench challenge: A machine learning contest for gene
  network inference from single-cell perturbation data}.
\newblock \emph{arXiv preprint arXiv:2308.15395}, 2023.

\bibitem[Kalatharan et~al.(2016)Kalatharan, Pei, Clemens, McTavish, Dixon,
  Rochon, Nash, Jain, Sarma, Zaleski, et~al.]{kalatharan2016positive}
Vinusha Kalatharan, York Pei, Kristin~K Clemens, Rebecca~K McTavish,
  Stephanie~N Dixon, Matthew Rochon, Danielle~M Nash, Arsh Jain, Sisira Sarma,
  Andrew Zaleski, et~al.
\newblock Positive predictive values of international classification of
  diseases, 10th revision coding algorithms to identify patients with autosomal
  dominant polycystic kidney disease.
\newblock \emph{Canadian Journal of Kidney Health and Disease}, 3:\penalty0
  2054358116679130, 2016.

\bibitem[Hern{\'a}n and Robins(2016)]{hernan2016using}
Miguel~A Hern{\'a}n and James~M Robins.
\newblock Using big data to emulate a target trial when a randomized trial is
  not available.
\newblock \emph{American journal of epidemiology}, 183\penalty0 (8):\penalty0
  758--764, 2016.

\bibitem[Abadie and Imbens(2016)]{abadie2016matching}
Alberto Abadie and Guido~W Imbens.
\newblock Matching on the estimated propensity score.
\newblock \emph{Econometrica}, 84\penalty0 (2):\penalty0 781--807, 2016.

\bibitem[Chen et~al.(2015)Chen, He, Benesty, Khotilovich, Tang, Cho, Chen,
  Mitchell, Cano, Zhou, et~al.]{chen2015xgboost}
Tianqi Chen, Tong He, Michael Benesty, Vadim Khotilovich, Yuan Tang, Hyunsu
  Cho, Kailong Chen, Rory Mitchell, Ignacio Cano, Tianyi Zhou, et~al.
\newblock Xgboost: extreme gradient boosting.
\newblock \emph{R package version 0.4-2}, 1\penalty0 (4):\penalty0 1--4, 2015.

\bibitem[Colosimo et~al.(2002)Colosimo, Ferreira, Oliveira, and
  Sousa]{colosimo2002empirical}
Enrico Colosimo, Fla{\'{}}~vio Ferreira, Maristela Oliveira, and Cleide Sousa.
\newblock Empirical comparisons between kaplan-meier and nelson-aalen survival
  function estimators.
\newblock \emph{Journal of Statistical Computation and Simulation}, 72\penalty0
  (4):\penalty0 299--308, 2002.

\bibitem[Davidson-Pilon(2019)]{davidson2019lifelines}
Cameron Davidson-Pilon.
\newblock lifelines: survival analysis in python.
\newblock \emph{Journal of Open Source Software}, 4\penalty0 (40):\penalty0
  1317, 2019.

\bibitem[Friberg et~al.(2018)Friberg, Gasparini, and
  Carrero]{friberg2018scheme}
Leif Friberg, Alessandro Gasparini, and Juan~Jesus Carrero.
\newblock A scheme based on icd-10 diagnoses and drug prescriptions to stage
  chronic kidney disease severity in healthcare administrative records.
\newblock \emph{Clinical Kidney Journal}, 11\penalty0 (2):\penalty0 254--258,
  2018.

\bibitem[Gimpel et~al.(2019)Gimpel, Bergmann, Bockenhauer, Breysem,
  Cadnapaphornchai, Cetiner, Dudley, Emma, Konrad, Harris,
  et~al.]{gimpel2019international}
Charlotte Gimpel, Carsten Bergmann, Detlef Bockenhauer, Luc Breysem, Melissa~A
  Cadnapaphornchai, Metin Cetiner, Jan Dudley, Francesco Emma, Martin Konrad,
  Tess Harris, et~al.
\newblock International consensus statement on the diagnosis and management of
  autosomal dominant polycystic kidney disease in children and young people.
\newblock \emph{Nature Reviews Nephrology}, 15\penalty0 (11):\penalty0
  713--726, 2019.

\bibitem[Wang et~al.(2022)Wang, Jiang, Thao, Sussman, LaBranche, Palmer,
  Harris, McKnight, Hoeflich, Schalm, et~al.]{wang2022protein}
Xiaofang Wang, Li~Jiang, Ka~Thao, Caroline~R Sussman, Timothy LaBranche,
  Michael Palmer, Peter~C Harris, G~Stanley McKnight, Klaus~P Hoeflich,
  Stefanie Schalm, et~al.
\newblock Protein kinase a downregulation delays the development and
  progression of polycystic kidney disease.
\newblock \emph{Journal of the American Society of Nephrology}, 33\penalty0
  (6):\penalty0 1087--1104, 2022.

\bibitem[Pratapa et~al.(2020)Pratapa, Jalihal, Law, Bharadwaj, and
  Murali]{pratapa2020benchmarking}
Aditya Pratapa, Amogh~P Jalihal, Jeffrey~N Law, Aditya Bharadwaj, and
  TM~Murali.
\newblock Benchmarking algorithms for gene regulatory network inference from
  single-cell transcriptomic data.
\newblock \emph{Nature methods}, 17\penalty0 (2):\penalty0 147--154, 2020.

\bibitem[Fr{\"o}hlich et~al.(2018)Fr{\"o}hlich, Kessler, Weindl, Shadrin,
  Schmiester, Hache, Muradyan, Sch{\"u}tte, Lim, Heinig,
  et~al.]{frohlich2018efficient}
Fabian Fr{\"o}hlich, Thomas Kessler, Daniel Weindl, Alexey Shadrin, Leonard
  Schmiester, Hendrik Hache, Artur Muradyan, Moritz Sch{\"u}tte, Ji-Hyun Lim,
  Matthias Heinig, et~al.
\newblock Efficient parameter estimation enables the prediction of drug
  response using a mechanistic pan-cancer pathway model.
\newblock \emph{Cell systems}, 7\penalty0 (6):\penalty0 567--579, 2018.

\bibitem[Dixit et~al.(2016)Dixit, Parnas, Li, Chen, Fulco, Jerby-Arnon,
  Marjanovic, Dionne, Burks, Raychowdhury, et~al.]{dixit2016perturb}
Atray Dixit, Oren Parnas, Biyu Li, Jenny Chen, Charles~P Fulco, Livnat
  Jerby-Arnon, Nemanja~D Marjanovic, Danielle Dionne, Tyler Burks, Raktima
  Raychowdhury, et~al.
\newblock {Perturb-Seq: dissecting molecular circuits with scalable single-cell
  RNA profiling of pooled genetic screens}.
\newblock \emph{cell}, 167\penalty0 (7):\penalty0 1853--1866, 2016.

\bibitem[Lopez et~al.(2020)Lopez, Gayoso, and Yosef]{lopez2020enhancing}
Romain Lopez, Adam Gayoso, and Nir Yosef.
\newblock Enhancing scientific discoveries in molecular biology with deep
  generative models.
\newblock \emph{Molecular Systems Biology}, 16\penalty0 (9):\penalty0 e9198,
  2020.

\bibitem[Consortium(2004)]{gene2004gene}
Gene~Ontology Consortium.
\newblock {The Gene Ontology (GO) database and informatics resource}.
\newblock \emph{Nucleic acids research}, 32\penalty0 (suppl\_1):\penalty0
  D258--D261, 2004.

\bibitem[Vaswani et~al.(2017)Vaswani, Shazeer, Parmar, Uszkoreit, Jones, Gomez,
  Kaiser, and Polosukhin]{vaswani2017attention}
Ashish Vaswani, Noam Shazeer, Niki Parmar, Jakob Uszkoreit, Llion Jones,
  Aidan~N Gomez, {\L}ukasz Kaiser, and Illia Polosukhin.
\newblock Attention is all you need.
\newblock \emph{Advances in neural information processing systems}, 30, 2017.

\bibitem[Dong et~al.(2023)Dong, Wang, Wei, de~O.~Fonseca, Perry, Frey, Ouerghi,
  Foxman, Ishizuka, Dhodapkar, et~al.]{dong2023causal}
Mingze Dong, Bao Wang, Jessica Wei, Antonio~H de~O.~Fonseca, Curtis~J Perry,
  Alexander Frey, Feriel Ouerghi, Ellen~F Foxman, Jeffrey~J Ishizuka, Rahul~M
  Dhodapkar, et~al.
\newblock Causal identification of single-cell experimental perturbation
  effects with cinema-ot.
\newblock \emph{Nature Methods}, 20\penalty0 (11):\penalty0 1769--1779, 2023.

\end{thebibliography}

\newpage
\section{Materials and methods}
\label{sec:methods}

\subsection{Problem setting}
\label{sec:problem}

We consider every experimental system subject to a perturbation (represented symbolically) for which we observe a readout. For example, an experiment could be conducted in a single-cell in vitro system in which transcript counts are measured after CRISPRi targeting a specific gene. A biological model system is considered to be a black box and no prior knowledge is assumed about the internal mechanism that gives rise to observed readouts.

\paragraph{Context (C).} The totality of the experimental context, including model system under study and the experimental protocol used, is represented by the variable $C\in\mathcal{C}$ and is referred to as the {context} of the experiment. The context $C$ is a symbolic description of the system itself and implicitly represents all the covariates that constitute the experimental conditions, for example, biological context details such as cell type, genetic background, and incubation protocols.

\paragraph{Perturbation (P).} We consider a {perturbation} any input to the system which is not already included in the context. A chemical compound, a gene knockout, or a disease that has perturbed the system are examples of perturbations. Let $P\in\mathcal{P}$ be the vector that describes a perturbation. Similar to the context $C$, $P$ is a symbolic representation of the perturbation. For instance, \texttt{CRISPRi\_STAT1} would symbolically represent CRISPR interference of gene STAT1.
Additionally, multi-perturbations that are symbolically represented as, e.g., \texttt{CRISPRi\_STAT1+CRISPRa\_FOXF1} (CRISPR interference of gene STAT1 coupled with CRISPR-mediated transcriptional activation of FOXF1), are modeled as a function of corresponding embeddings. In the experiments in this paper, we used embedding average.

\paragraph{Readout (R).} The symbolic description of the measurements observed in the system which is under perturbation is represented by a {readout} $R\in\mathcal{R}$ where $\mathcal{R}$ is a set of symbols that correspond to all possible discrete values that represent observed readouts. For example, $R$ can represent the gene expression of the gene PSMA1, denoted as \texttt{Transcript\_PSMA1}.

\paragraph{Predicted experimental observation (Y).} The concrete measurement taken in context $C$ after perturbation $P$ using readout $R$ is represented by $Y\in\mathcal{Y}\subseteq\mathbb{R}$.  It is of note that the experimental observation $Y$ is distinct from the readout $R$ in that $R$ symbolically describes the type of measurement taken whereas $Y$ is a concrete instance of that measurement in the experimental context $C$ under perturbation $P$. \\

Let $O = (P, R, C, Y)$ be the stack of aforementioned random variables and $\mathcal{I}=\{1,2, \ldots\}$ be the index set of all possible potential observed samples. Therefore, the index $i\in\mathcal{I}$ refers to one potential observation $O^{(i)} = (P^{(i)}, R^{(i)}, C^{(i)}, Y^{(i)})$. Let $\mathcal{D}_\textrm{obs} = \{O^{(1)}, O^{(2)}, \ldots, O^{(n_\textrm{obs})}\}$ be the set of observations which has $n_\textrm{obs}$ data points and $\mathcal{I}_\textrm{obs}\subseteq \mathcal{I}$ be the set of associated indices. It is clear that $Y$ is not independent from $P$, $R$, and $C$ . We want to learn the causal model $q(Y|do(P=p), R, C)$. Here, $q$ is the probability distribution of the outcome $Y$ in a biological system within the context $C$ when the perturbation $p$ is applied, and the readout $R$ is observed. We would like to leverage the structural dependence between these variables to estimate $q$ from $\mathcal{I}_\textrm{obs}$ so it can predict the outcome of unobserved (perturbation, readout, context) combinations indexed by $j\in \mathcal{I}_\textrm{unobs} = \mathcal{I} \backslash \mathcal{I}_\textrm{obs}$. Mathematically, we want to estimate

\begin{equation}
\label{eq:counterfactual_outcome_from_original_space}
    q(Y | P, R, C, \mathcal{I}_\textrm{obs})
\end{equation}
for any combination $(P, R, C)\in \mathcal{P}\times \mathcal{R}\times \mathcal{C}$. This is only possible if the spaces $\mathcal{P}$, $\mathcal{R}$, and $\mathcal{C}$ have some structure that allows defining the concept of distance. For example, for a system with context $C^{(j)}$, predicting the effect of perturbation $P^{(j)}$ on readout $R^{(j)}$ is possible if the outcome of a {similar} perturbation on a {similar} readout is already observed for a system within a {similar} context. Clearly, talking about similarities requires the concerned spaces to possess some structure in which a distance metric can be defined. As $(P, R, C)$ are in essence discrete symbolic values, it is necessary to first transform them into more tractable spaces that we call {embedding spaces}.  Let  $Z_p\in\mathcal{Z}_p\subseteq\mathbb{R}^{d_{z_p}}$, $Z_r\in\mathcal{Z}_r\subseteq\mathbb{R}^{d_{z_r}}$, and $Z_c\in\mathcal{Z}_c\subseteq\mathbb{R}^{d_{z_c}}$, be the random variables that represent the embeddings of $P$, $R$, and $C$ respectively. The transformation maps $\phi_{p}:\mathcal{P}\to\mathcal{Z}_p$ $\phi_{r}:\mathcal{R}\to\mathcal{Z}_r$, and $\phi_{c}:\mathcal{C}\to\mathcal{Z}_c$ that induce such structure in the embedding spaces are learned from $\mathcal{I}_\textrm{obs}$. In other words, the information of the observed data is learned in $\phi_p(\cdot)$, $\phi_r(\cdot)$, and $\phi_c(\cdot)$ functions. This means that for any unseen $(P=p,R=r,C=c)$ tuples, their corresponding embeddings $Z_p$,  $Z_r$, and $Z_c$ implicitly contain some information from $\mathcal{I}_\textrm{obs}$. This is indeed the reason that enables knowledge transfer to unseen perturbations, readouts and contexts. With the learned embedding space, \cref{eq:counterfactual_outcome_from_original_space} can be written as

\begin{equation}
    \label{eq:counterfactual_outcome_from_embedding_space}
    q_\textrm{emb}(Y | Z_p, Z_r, Z_c, \mathcal{I}_\textrm{obs})
\end{equation}
where the subscript \enquote{emb} emphasized that the map is defined from the embedding spaces instead of the original spaces. Due to the learned structure in the embedding spaces, it is expected that $q_\textrm{emb}(\cdot)$ be more accessible to learn than $q(\cdot)$.

\subsection{Model details}
 \label{sec:model}

\begin{figure*}[!t]
\centering
\includegraphics[width=\textwidth]{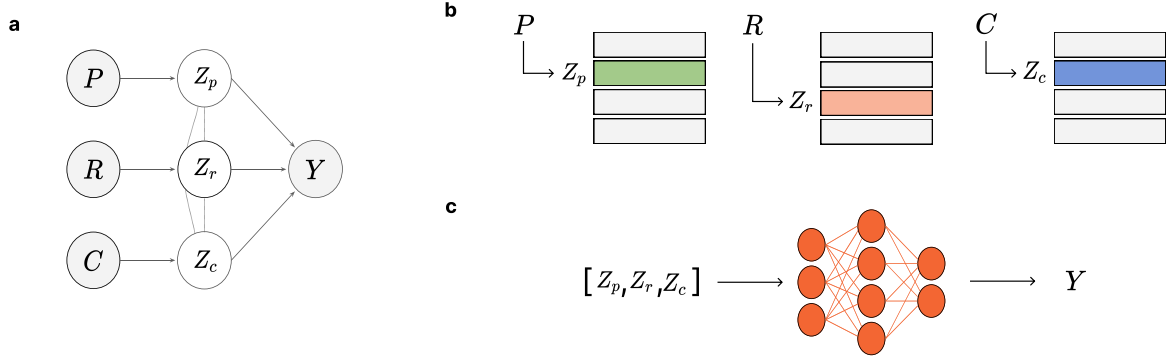}
\caption{\textbf{Model description.} \textbf{a.} Graphical model shows the dependencies between random variables previously described in \cref{sec:problem}. Dashed lines indicate implicit bi-directional dependencies that enable transfer learning across datasets. Symbolic perturbation, readout, and context descriptors ($P$,$R$,$C$) are first embedded ($Z_p$, $Z_r$, $Z_c$), then used to generate output $Y$ that represents the value of the readout $R$. \textbf{b.} Embeddings are implemented as learnable look-up tables. $P$, $R$, and $C$ identify indices in the corresponding tables. \textbf{c.} Concatenated embeddings are forward propagated through a multilayer perceptron to predict the output $Y$.}
\label{fig:model}
\end{figure*}

Building on the problem formulation described in \cref{sec:problem}, we designed the architecture of LPM as shown in \cref{fig:model}a. Since $P$, $R$ and $C$, are discrete random variables, it is simple to implement the corresponding embeddings $Z_p$, $Z_r$, $Z_c$ using symbol vocabularies and learnable look-up tables (see \cref{fig:model}b). Symbol vocabularies map symbols to indices, while look-up-tables map indices to learnable weights that we treat as embeddings. This model can nevertheless be trivially generalized to include more complex perturbations or context descriptions, e.g., multiple perturbations can be implemented as a sum of individual perturbations \cite{roohani2023predicting,lotfollahi2023predicting}. The prediction network is a neural network that is learned end-to-end together with the embeddings, by backpropagating the error. We found the multilayer perceptron (MLP) architecture with ReLU activation functions, implemented on top of concatenated embeddings, to work satisfactorily (see \cref{fig:model}b). We note that an extensive architecture search was not performed and further architecture tuning could potentially further improve results.

The key property of our model which enables scaling training across heterogeneous high-throughput perturbation screens (\cref{fig:intro}), is conditioning on the readout $R$. To clarify why this simple trick is effective, consider an alternative description of the causal model \cref{eq:counterfactual_outcome_from_original_space} that does not condition on the $R$, i.e., $q\prime(Y|P,C, \mathcal{I}_\textrm{obs})$ \citep{wu2022variational}. In this case, $Y\in\mathcal{Y}\subseteq\mathbb{R}^{d_y}$ is a vector (not a scalar) whose dimension $d_y$ is the number of readouts. The challenge is that \emph{de facto} each perturbation screen has its own subset of phenotypic readouts. Even if the same modality is measured in two datasets, e.g. transcriptome, datasets often measure different parts of the transcriptome. The problem exacerbates if different modalities are used for training (e.g., proteome along with transcriptome), or if a large number of datasets is included in the training process. Related previous works alleviate this issue by selecting only readouts that appear in all considered perturbation experiments. However, this is clearly suboptimal as we discard relevant information, and furthermore, this strategy becomes inapplicable if we want to scale across a large number of datasets as the size of the overlapping set only shrinks as the number of datasets increases. Moreover, certain experimental measurement technologies, such as DRUG-seq \citep{ye2018drug} may contain missing values. \themethod{} is designed to be robust to missing readouts as well.

\subsection{Data description}
\label{sec:data-description}

In our first set of experiments, we used data from both single-cell and bulk experiments to demonstrate the flexibility of the design of \themethod{}. Single cell data contains two experimental contexts from \citep{replogle2022mapping}, \reploglek{} and \reploglerpe{}. The data is based on transcriptome measurements generated after DepMap essential genes \cite{tsherniak2017defining} have been perturbed using CRISPRi Perturb-seq technology.  The data were sequenced in chronic myeloid leukaemia (K562) and retinal pigment epithelial (RPE1) cell lines respectively. A z-normalized version of the datasets was used as recommended and provided by the authors. Z-normalization was performed per gemgroup (batch). Single-guide RNAs (sgRNAs) that target the same gene were aggregated to represent a single perturbation. We removed cells containing multiple knockdowns to simplify the evaluation, focusing exclusively on predicting unobserved perturbations rather than combinations of observed perturbations. The second line of experimental contexts was obtained from the Connectivity Map (CMAP) project. In particular, we used the expanded Connectivity Map Lincs 2020 screens (\url{https://clue.io}) \citep{subramanian2017next}, both on pharmacological and CRISPR knockout (CRISPR-KO) perturbations. There were overall 26 biological contexts used from the LINCS studies based on bulk data, including different cell types but also two different types of perturbations (CRISPR-KO and compound-based perturbations). We discarded LINCS contexts that had too few perturbations (<300), for simplicity, since they did not make a difference in our analysis. To further simplify the analysis, we used only most-commonly appearing drug doses (10$\mu M$) and observation times (24h). \rnn{We used the preprocessed data that included quality control as described by \citet{replogle2022mapping} and \citet{subramanian2017next}.}

\rnn{In our second set of experiments, we used data from both single-perturbation experiments \reploglek{} and \reploglerpe{}, as well as multi-perturbation experiments from Norman et al. \citep{norman2019exploring}.
The data was processed as described in \cite{roohani2023predicting}, log-transformed and filtered to 5000 highly-variable genes.}
The overview of all used data is given in \cref{tab:datasets}.

\begin{table*}[h!]
    \centering
    \setlength{\tabcolsep}{4pt}
    \begin{tabular}{lrrrr}
        \toprule
         Biological context\hspace{5em} & Perturbation & \hspace{3em}\#Cells & \#Readouts & \#Perturbations \\
        (C) & type & (N) & (R) & (P)\\
        \midrule
        \reploglek{} \citep{replogle2022mapping} & CRISPRi & \numprint{310385}& \numprint{8562} & \numprint{2056} \\
        \reploglerpe{} \citep{replogle2022mapping} & CRISPRi & \numprint{247914}& \numprint{8748} & \numprint{2392} \\
       LINCS (A375) \citep{subramanian2017next} & CRISPR-KO & n/a - bulk& \numprint{978} & \numprint{1904} \\
       LINCS (A549) \citep{subramanian2017next} & CRISPR-KO & n/a - bulk& \numprint{978} & \numprint{1917} \\
       LINCS (AGS) \citep{subramanian2017next} & CRISPR-KO & n/a - bulk& \numprint{978} & \numprint{1464} \\
       LINCS (BICR6) \citep{subramanian2017next} & CRISPR-KO & n/a - bulk& \numprint{978} & \numprint{1659} \\
       LINCS (ES2) \citep{subramanian2017next} & CRISPR-KO & n/a - bulk& \numprint{978} & \numprint{2117} \\
       LINCS (HT29) \citep{subramanian2017next} & CRISPR-KO & n/a - bulk& \numprint{978} & \numprint{1385} \\
       LINCS (MCF7) \citep{subramanian2017next} & CRISPR-KO & n/a - bulk& \numprint{978} & \numprint{1049} \\
       LINCS (PC3) \citep{subramanian2017next} & CRISPR-KO & n/a - bulk& \numprint{978} & \numprint{772} \\
       LINCS (U251MG) \citep{subramanian2017next} & CRISPR-KO & n/a - bulk& \numprint{978} & \numprint{2419} \\
       LINCS (YAPC) \citep{subramanian2017next} & CRISPR-KO & n/a - bulk& \numprint{978} & \numprint{994} \\
       LINCS (A375) \citep{subramanian2017next} & Compounds & n/a - bulk& \numprint{978} & \numprint{1803} \\
       LINCS (A549) \citep{subramanian2017next} & Compounds & n/a - bulk& \numprint{978} & \numprint{1484} \\
       LINCS (ASC) \citep{subramanian2017next} & Compounds & n/a - bulk& \numprint{978} & \numprint{391} \\
       LINCS (HA1E) \citep{subramanian2017next} & Compounds & n/a - bulk& \numprint{978} & \numprint{1463} \\
       LINCS (HCC515) \citep{subramanian2017next} & Compounds & n/a - bulk& \numprint{978} & \numprint{1067} \\
       LINCS (HELA) \citep{subramanian2017next} & Compounds & n/a - bulk& \numprint{978} & \numprint{492} \\
       LINCS (HEPG2) \citep{subramanian2017next} & Compounds & n/a - bulk& \numprint{978} & \numprint{598} \\
       LINCS (HT29) \citep{subramanian2017next} & Compounds & n/a - bulk& \numprint{978} & \numprint{1082} \\
       LINCS (MCF10A) \citep{subramanian2017next} & Compounds & n/a - bulk& \numprint{978} & \numprint{524} \\
       LINCS (MCF7) \citep{subramanian2017next} & Compounds & n/a - bulk& \numprint{978} & \numprint{2035} \\
       LINCS (MDAMB231) \citep{subramanian2017next} & Compounds & n/a - bulk& \numprint{978} & \numprint{384} \\
       LINCS (NPC) \citep{subramanian2017next} & Compounds & n/a - bulk& \numprint{978} & \numprint{482} \\
       LINCS (PC3) \citep{subramanian2017next} & Compounds & n/a - bulk& \numprint{978} & \numprint{2307} \\
       LINCS (THP1) \citep{subramanian2017next} & Compounds & n/a - bulk& \numprint{978} & \numprint{342} \\
       LINCS (VCAP) \citep{subramanian2017next} & Compounds & n/a - bulk& \numprint{978} & \numprint{938} \\
       LINCS (YAPC) \citep{subramanian2017next} & Compounds & n/a - bulk& \numprint{978} & \numprint{414} \\
        \bottomrule
    \end{tabular}
    \caption{\textbf{Data overview.} The names in the brackets of experimental contexts indicate cell types. Perturbation counts were computed after the cells that were perturbed with multiple gRNAs targeting different genes were removed. All LINCS contexts have 978 readouts (L1000 transcriptome).}
    \label{tab:datasets}
\end{table*}%

\subsection{Experimental setup}
\label{sec:exp-setup}

\paragraph{Perturbation prediction.} We evaluated how well \themethod{} performs in terms of predicting expected response to different "unobserved" perturbations (\cref{fig:intro}b). For datasets with single-cell readouts, \themethod{} was trained on mean-aggregated data to reflect effects across cell populations. We randomly selected 8 biological contexts, covering both bulk and single-cell data, as well as pharmacological and genetic interventions. Perturbation data in each considered biological context (\cref{tab:datasets}) were split into 70\%, 15\% and 15\% folds for model training, validation and testing, respectively. We performed cross-validation, where in each fold, the test data from one of the considered biological contexts was isolated and the rest of the data (all the data from \cref{tab:datasets}) was used for training the \themethod{}. For baselines with predefined embeddings, we used only the training data from the target biological context (including other data was not found to affect performance). Validation sets were used for hyper-parameter selection and early stopping.

Model predictions were evaluated against the ground truth with respect to: {(i)} Root mean squared error ({RMSE}), {(ii)} the coefficient of determination ($R^2$), {(iii)} Pearson correlation coefficient ({Pearson}), and {(iv)} Mean absolute error ({MAE}). Additionally, we evaluated performance on different portions of test data. We have included the full test set but also 25\% and 10\% portions of the test data that contained the strongest perturbations and most moving readouts. The magnitude of perturbations and the magnitude of readout movements were measured with respect to RMSE against mean readout values in the control data. The following competing methods were examined:
\begin{itemize}[leftmargin=0.5cm]
    \item \textbf{NoPerturb:} This baseline is similar to the one described by \cite{roohani2023predicting}. When predicting the outcome of a perturbation, it takes no perturbation information into account but assigns each readout the average value as observed in the training data.
    \item \textbf{Catboost:} We used Catboost algorithm \cite{prokhorenkova2018catboost} in combination with different predefined embeddings (explained in more detail in \cref{sec:other-related-work}). For genetic perturbations, we found that multi-hot embeddings of the Reactome database \cite{fabregat2018reactome} worked the best. Reactome is an open-source, open-access, manually curated and peer-reviewed pathway database. Essentially, for each perturbed gene, we created a multi-hot vector depending on whether it belongs to a certain Reactome pathway or not. We then trained a CatBoost to predict the expected response to a perturbation based on Reactome embeddings.
    \item \textbf{Geneformer:} We used the pretrained Geneformer model published at \url{https://huggingface.co/ctheodoris/Geneformer} (downloaded in August 2023). We either fine-tuned the model according to their respective instructions or used their embeddings with a CatBoost model, then chose the approach that performed better. The source Geneformer model was pretrained on a large corpus of unperturbed pooled single-cell gene expression datasets. We note that Geneformer is limited to processing at most 2048 transcripts, we therefore created multiple references that were independently fed to Geneformer to produce gene embeddings for all perturbed genes across the evaluated datasets. Geneformer is described in detail in \cite{theodoris2023transfer}.
    \item \textbf{scGPT:} We used the pretrained whole-human scGPT model published at \url{https://github.com/bowang-lab/scGPT} (downloaded in August 2023). We either fine-tuned the model according to their respective instructions or used their embeddings with a CatBoost model, then chose the approach that performed better. scGPT is trained on pooled single-cell gene expression data not under perturbations. scGPT is described in detail in \cite{cui2023scgpt}.
    \item \textbf{GenePT:} We followed the strategy outlined in \citet{chen2023genept} to generate gene embeddings from the National Center for Biotechnology Information (NCBI) summary text descriptions of genes using ChatGPT 3.5, and then used those gene embeddings as the input data from which a Catboost model was trained to make downstream predictions for evaluation. GenePT is described in detail in \cite{chen2023genept}.
    \item \textbf{GEARS:} The model covered in Section \ref{sec:other-related-work} and in detail described in \cite{roohani2023predicting}.
    \item \textbf{CPA:} The model covered in Section \ref{sec:other-related-work} and in detail described in \cite{lotfollahi2023predicting}.
\end{itemize}

Apart from weighted Adam \cite{kingma2014adam} optimizer and early stopping (patience of 10 epochs) that were fixed throughout our study, the rest of the hyper-parameters were tuned as indicated in \cref{tab:hps}. The baselines were given the same computational budget and were tuned according to the recommendations from the corresponding publications. It took about 50 epochs for all training runs to early stop based on the validation performance on the validation set. Early stopping used RMSE as the stopping criterion.

\begin{table}[!t]
    {\scriptsize
        \centering
        \begin{tabular}{l|llllllll}
            \toprule
                     & Learning & Learning & MLP & Dropout  & Hidden & Embed. & Batch \\
                     & rate & rate decay & layers & rate & dim. &  dim. & size \\
            \toprule
            Grid search & 0.001 & 0.99 & 1 & 0 & 256 & 128 & 1000 \\
                         & 0.002 & 0.97 & 2 & 0.1 & 512 & 64 & 5000 \\
                         & 0.01 & &  & 0.25 & & 32 & 10000\\
                         & &  &  &  & & \\
            \midrule
            Target context & & & & & & &\\
            Replogle (K562) CRISPRi & 0.002 & 0.99 & 2 & 0 & 512 & 32 & 5000\\
            Replogle (RPE1) CRISPRi & 0.005 & 0.99 & 2 & 0 & 512 & 32 & 5000 \\
            LINCS (HT29) Compounds & 0.002 & 0.97 & 2 & 0.1 & 256 & 128 & 1000\\
            LINCS (HELA) Compounds\hspace{3.75em}  & 0.002 & 0.97 & 2 & 0.1 & 256 & 128 & 1000\\
            LINCS (HA1E) Compounds & 0.002 & 0.97 & 2 & 0.1 & 256 & 128 & 1000\\
            LINCS (MCF7) CRISPR-KO & 0.002 & 0.97 & 2 & 0.1 & 256 & 128 & 1000\\
            LINCS (HT29) CRISPR-KO & 0.002 & 0.97 & 2 & 0.1 & 256 & 128 & 1000\\
            LINCS (A549) CRISPR-KO & 0.002 & 0.97 & 2 & 0.1 & 256 & 128 & 1000\\
            \midrule
            Training contexts & & & & & & &\\
            Replogle all & 0.002 & 0.99 & 2 & 0 & 512 & 32 & 5000\\
            LINCS all & 0.002 & 0.97 & 2 & 0.1 & 256 & 128 & 1000\\
          \bottomrule
        \end{tabular}
        \caption{\textbf{Hyper-parameter selection.} The table consists of three parts. In the first part shown are the hyper-parameters that were used in the grid search as a part of the model selection. The second part shows the selected hyper-parameters for the post-perturbation transcriptome prediction benchmark. The third part shows hyper-parameters used for the embedding-related tasks, where two \themethod{}s were trained on all LINCS and Replogle data.\label{tab:hps}}
    }
\end{table}

\paragraph{Embedding evaluation.} For embeddings-related experiments, we used consensus hyper-parameter configuration from our predictive performance experiments. The perturbation embeddings from the trained models were first used to assess the information content by classifying molecular functions associated with the perturbations. For that purpose, we trained CatBoost models with 5 different random seeds to classify perturbations into functional mechanisms as annotated by \citet{replogle2022mapping} based on the perturbed genes' mechanisms using \themethod{} embeddings and various state-of-the-art gene embeddings (Reactome, STRING, Achilles, Gene2vec, scGPT, Geneformer and GenePT) (\cref{fig:lincs-embeddings}). We then compared the ability of the compared embeddings to accurately assign functional mechanisms to genes using the Area under the Curve (AUC) using the unweighted mean in a one-vs-rest multi-class computation to aggregate across the functional categories.
To evaluate how well the trained \themethod{} can identify mechanisms of action of the compound inhibitors, we used drug repurposing hub data (\url{https://repo-hub.broadinstitute.org/repurposing}). We kept only targets and compounds that appear in our data (from \cref{tab:datasets}), and only information about compounds that inhibit/suppress the targets (removing compounds with non-inhibiting mechanisms).

\paragraph{Gene-gene network inference evaluation.} To learn gene-gene interaction networks, we used the state-of-the-art method proposed by \citet{deng2023supervised} (Guanlab) complemented with data imputed by an {\themethod}. Guanlab is generally biased towards predicting edges for the nodes that have been perturbed in the training data, meaning that almost no edges are predicted from non-perturbed nodes. This impacts the recovery rate of causal gene interactions (as measured by the FOR) \cite{chevalley2023causalbench}. To remedy this and to leverage the {\themethod} perturbation predictions, we implemented a two-step approach. The first step involved running the Guanlab algorithm on the original training data, and taking the top $2500$ predicted edges. We then retrained a Guanlab model, this time only on the prediction from the {\themethod}, taking the top $2500$ predicted edges. The final output consisted of the union of those two sets of $2500$ edges. Compared to the original Guanlab (top 5k) model without imputation, we obtained a significantly lower FOR for the same number of output edges, indicating that leveraging {\themethod}s allowed recovering many more genetic interactions.

\paragraph{In-silico perturbation study of predicted PKD1 upregulators.} We trained an \themethod{} on pooled chemical and genetic perturbation data from LINCS and, after training, evaluated the set of PRC queries (P = all \numprint{5310} chemical perturbations in LINCS, R = all \numprint{978} L1000 transcripts, C = HA1E LINCS). Because PKD1 was not in the set of landmark transcripts directly experimentally measured in LINCS, we additionally trained a Catboost \cite{prokhorenkova2018catboost} regressor model (hyperparameters: 512 iterations, 100 round early stopping patience) using HA1E cells to infer the PKD1 value corresponding to the L1000 transcripts predicted by \themethod{}. This produced an in-silico predicted PKD1 value for each of the \numprint{5310} chemical perturbation included in LINCS. We then ranked the chemical perturbations from highest predicted PKD1 upregulation to lowest and filtered for clinical stage drugs. We removed natural products and early stage investigational drugs from the ranking because it is a priori unclear whether their safety profile or off-target effects may impede clinical utility as potential ADPKD therapeutics. We note that the ability of \themethod{} to perform context-specific predictions was essential for the study results as predictions in other cellular model systems did not surface the same chemical perturbations as top regulators of PKD1 - indicating a dependence of the results of the in-silico perturbation experiment on the biological context in kidney cells.

\begin{table}[!t]
    {\scriptsize
        \centering
\begin{tabular}{p{12.5em}|rrrr}%
\toprule%
& \multicolumn{2}{c}{Before matching}& \multicolumn{2}{c}{After matching}\\
Property&Simvastatin &No statin use&Simvastatin &No statin use\\%
\midrule%
Patients&\numprint{1716}&\numprint{48555}&\numprint{1594}&\numprint{1594}\\%
ADPKD&100\%&100\%&100\%&100\%\\%
Age&63.75&52.41&63.40&64.72\\%
95\% CI&(41.10, 79.58)&(14.33, 79.84)&(41.34, 79.77)&(42.34, 79.21)\\%
Index year&2012.94&2016.73&2013.23&2012.42\\%
95\% CI&(2008.85, 2018.49)&(2010.31, 2021.83)&(2009.12, 2018.64)&(2008.85, 2017.90)\\%
Sex (Female)&42.31\%&52.08\%&43.41\%&42.10\%\\%
Ethnicity&&\\%
Caucasian&85.02\%&73.95\%&84.32\%&85.57\%\\%
African American&9.91\%&12.40\%&10.54\%&9.66\%\\%
Other/Unknown&3.55\%&11.39\%&3.64\%&3.14\%\\%
Asian&1.52\%&2.26\%&1.51\%&1.63\%\\%
Hispanic&3.26\%&6.56\%&3.14\%&3.01\%\\%
Region&&\\%
Region Midwest&61.83\%&41.94\%&61.98\%&66.50\%\\%
Region South&13.05\%&23.98\%&13.93\%&9.60\%\\%
Region Northeast&11.83\%&15.21\%&11.04\%&10.92\%\\%
Region West&10.31\%&14.06\%&10.10\%&10.54\%\\%
Region Other/Unknown&2.97\%&4.82\%&2.95\%&2.45\%\\%
Type 1 diabetes&4.66\%&1.87\%&3.70\%&5.02\%\\%
Type 2 diabetes&42.60\%&19.04\%&40.09\%&45.29\%\\%
Hypertension&94.52\%&65.20\%&94.17\%&95.73\%\\%
Cardiovascular disease&56.00\%&28.21\%&54.02\%&59.47\%\\%
Obesity&43.88\%&23.65\%&41.84\%&46.68\%\\%
Proteinuria&27.10\%&11.53\%&24.47\%&28.48\%\\%
          \bottomrule
\end{tabular}%
        \caption{\textbf{Retrospective study cohort descriptions.} Cohort descriptions for the retrospective matched cohort study conducted to evaluate impact of Simvastatin on progression to ESRD among ADPKD diagnosed individuals (\cref{fig:discovery}). Each row corresponds to a covariate measured across the cohorts (Simvastatin 1 year+ exposure vs no statin use) at index date before and after (the two rightmost columns) propensity score matching. The cohorts included in the study are well balanced and comparable across the indicated covariates that include unspecific (e.g., age) and specific (e.g., diabetes and obesity) risk factors for ESRD.\label{tab:cohorts}}
    }
\end{table}

\paragraph{Retrospective matched cohort study to evaluate simvastatin in ADPKD.} To validate the predicted PKD1 upregulator simvastatin in real-world clinical data, we performed a matched cohort study using retrospective data collected in the Optum\textsuperscript{\textcopyright} de-identified Electronic Health Record database collected from a decentralised network of healthcare provider organisations, including \numprint{2000} hospitals, \numprint{7000} clinics and 100 million unique individuals, from 2007 to 2023 in the United States (US). Among those clinically diagnosed with ADPKD (International Classification of Diseases [ICD] 10: Q61.2 or Q61.3 \cite{kalatharan2016positive}; ICD 9: 753.12 or 753.13), we created two cohorts for comparison: a treatment cohort consisting of individuals that were exposed to Simvastatin for more than one year (date of first and last prescription greater than 365 days apart) and a \enquote{not exposed} cohort of individuals that did not receive any of the statins (simvastatin, fluvastatin, pitavastatin, cerivastatin, mevastatin) predicted by a \themethod{} to upregulate PKD1. For the treatment cohort, we used 1 year after their first recorded simvastatin prescription as the index date, i.e. there was a 1 year induction period for simvastatin to influence cystogenesis. For the \enquote{not exposed} cohort, we picked an index date between their first and last entry into the EHR at random (uniform distribution) in line with the target trial emulation framework \cite{hernan2016using}. Censoring was applied on the date of the last recorded entry for an individual in the EHR and the first recorded entry in the EHR was considered the first date of observation for each individual. We applied 1:1 propensity score matching (PSM) \cite{abadie2016matching} using a non-linear XGBoost \cite{chen2015xgboost} propensity score estimator with a 1\% calliper to ensure the two cohorts are comparable based on their covariates as measured prior to their respective index dates (see \cref{tab:cohorts} for matching covariates). This yielded two cohorts comparable in their observed covariates (\cref{tab:cohorts}) for which we used a Nelson-Aalen estimator \cite{colosimo2002empirical} using the \texttt{lifelines} package \cite{davidson2019lifelines} to produce cumulative hazard estimates for progression to ESRD (ICD 10: N18.* or N17 or N19 or Z99.2; ICD 9: 585.* or 586 or 584.9 or V45.11 \cite{friberg2018scheme}; * indicates inclusion of the primary code and all sub-codes). We chose the Nelson-Aalen estimator because it does not require the proportional hazards assumption. We note that we were not able to validate other predicted PKD1 upregulators in real-world clinical data because there are not enough individuals that received the other predicted medications and are diagnosed for ADPKD in the Optum\textsuperscript{\textcopyright} EHR database. We note that the observed association of long-term simvastatin exposure in reducing ESRD progression in ADPKD diagnosed individuals does not conclusively prove that PKD1 is the responsible mechanism as it is possible that simvastatin reduces ESRD events through other or multiple means.

\paragraph{Tolvaptan.} Tolvaptan is the only medicine currently approved for treatment of ADPKD \cite{gimpel2019international}. Tolvaptan is an AVPR2 antagonist that reportedly slows disease progression by inhibiting production of cyclic adenosine 3',5'-monophosphate (cAMP) \cite{wang2022protein}, and therefore a different mechanism than the mechanism targeting PKD1 upregulation that we investigated in the presented in-silico study. Because the training data used to develop \themethod{} did not contain reference data for tolvaptan perturbations, we predicted the effects of its analogue perturbation in silico via CRISPRi of AVPR2. We found that  the predicted impact of inhibiting AVPR2 on PKD1 was (although also directionally positive) minor compared to statins predicted to upregulate PKD1 (\Cref{fig:discovery}) - indicating that tolvaptan indeed likely acts through a different mechanism than PKD1 upregulation.

\section{Extended related work}
\label{sec:other-related-work}

In-silico perturbation modeling, in the sense of predicting post-perturbation outcome, has been studied intensively in recent years. A traditional approach is to first infer a perturbation screen-specific biological network and then use the inferred network to predict perturbation response with respect to outcomes of interest \citep{pratapa2020benchmarking,yuan2021cellbox,kamimoto2023dissecting}. For example, CellOracle \cite{kamimoto2023dissecting} infers a linear representation of a gene regulatory network (GRN) which is then used to predict gene expression and cell identity changes caused by hypothetical perturbations of transcription factors (TFs). This approach is more classical in the sense that it involves several complex steps (clustering, graph inference, prediction, dimensionality reduction), as opposed to deep learning-based approaches where models are trained "end-to-end". Other non-end-to-end approaches include mechanistic modeling of cell machinery via ODEs \citep{frohlich2018efficient,yuan2021cellbox} or using linear regression to describe perturbation-readout relationships \citep{dixit2016perturb}. Most of the above-mentioned methods are also limited in the type of perturbations they model (e.g., mutations of TFs) or the readouts they predict (e.g., cell viability or growth rate). In contrast, \themethod{} handles arbitrary types of perturbations or readouts. A notable advantage of traditional approaches is that they offer a certain degree of interpretability such as an explicit representation of a GRN.

More recently, deep learning methods have been popularized \citep{lopez2020enhancing} due to their ability to model functions of arbitrary complexity and leverage massive amounts of data from high-throughput screens. Variational autoencoder (VAE)-based methods \citep{lotfollahi2019scgen, hetzel2022predicting, lopez2022learning, wu2022predicting, wu2022variational, lotfollahi2023predicting} autoencode readouts, normally transcriptome, through a latent space in which perturbations are modeled and applied. VAE-based methods are rather flexible in terms of the types of perturbations and covariates they can deal with. For example, CPA \citep{lotfollahi2023predicting} incorporates drug dosage information and handles combinatorial perturbations such as multi-gene knockouts.
CPA and related methods learn perturbation and covariate embedding spaces end-to-end, in a fully data-driven fashion.
One disadvantage of VAE methods is that they require each perturbation to be experimentally perturbed
before the effect of perturbing the combination can be predicted, i.e., they are limited to \emph{in-vocabulary} settings.
Like in \themethod{}, this could be alleviated through data pooling across different studies, however, due to the rigid encoder-decoder architecture, VAEs are generally not suitable for handling heterogeneous screens that may differ in terms of data format and dimensions. Data pooling can be enabled using complex tricks that involve architectural changes that limit the size of the pool \citep{hetzel2022predicting}.

\rnn{To enable \emph{out-of-vocabulary} prediction, a line of methods that incorporate prior knowledge emerged \citep{wu2022predicting,hetzel2022predicting,roohani2023predicting}.
ChemCPA~\citep{hetzel2022predicting} uses molecular structures to construct drug embeddings and predicts cellular response to drugs that do not appear in the experimental data.
GEARS~\citep{roohani2023predicting} uses publicly available Gene Ontology (GO) and gene co-expression data to construct gene embeddings to extrapolate response prediction to unseen genetic perturbations.
However, predefined representations of perturbations have their own issues.
Firstly, the extrapolation power is limited by the coverage of those perturbations. For example, it is unclear how to obtain common perturbation joint embedding spaces of pharmacological and genetic perturbations, which could be very useful in identifying drug-gene relationships as shown in \cref{fig:lincs-embeddings}. Secondly, unlike representations learned end-to-end, predefined ones are created independently of the perturbation response prediction task hence may not be sufficiently informative of functional relationships among perturbations.
Another disadvantage of GEARS~\citep{roohani2023predicting} is that is not suitable for multi-context settings.}

Our work follows an end-to-end learning but is focus on integrating perturbation screens from different contexts.
To our knowledge, ChemCPA \cite{hetzel2022predicting} is the only method from the VAE space which attempts to deal with heterogeneity in multi-context settings. ChemCPA uses what they call \enquote{architecture surgery}, wrapping pre-trained VAE with supplementary encoders and decoders to adapt the model to the target data format. This strategy unfortunately (i) ignores the absence of missing readouts in the pre-training data; (ii) is applicable only when all the pre-training data is of equivalent format, otherwise, undesirable subsetting of readouts must be applied which discards relevant experimental information.
In contrast to ChemCPA, the proposed non-autoencoder architecture of an LPM can be trained on datasets that contain arbitrary types of perturbations, readouts, and contexts, without making assumptions about the dimensionality of the data.
Another major difference between VAE-based approaches such as CPA and \themethod{} is the structure that such methods impose on the latent space. For example, CPA takes a so-called \enquote{information removal} strategy which optimizes auxiliary loss functions to minimize the mutual information among embeddings encouraging them to be independent. This step is required due to the fact that the encoder can tangle the information content of the embeddings. However, \themethod{} is encoder-free and thus does not need to explicitly disentangle information from the output of the encoder. We note that \themethod{} achieves disentangled embeddings without explicitly optimizing for it (\cref{fig:lincs-embeddings}).

A critical component of \themethod{} and related works is the representation of perturbations in the form of embeddings. \themethod{} and CPA train embeddings from scratch, learning them from data in an unbiased fashion. On the other hand, GEARS and the Catboost baseline (from our benchmarks) utilise predefined embeddings. GEARS embeddings are based on Gene Ontology (GO) \cite{gene2004gene} and gene co-expression knowledge graph on top of which a graph neural network is applied. For the Catboost baseline, we tried multiple different ways to represent both readout and perturbation embeddings, including Gene2vec \cite{du2019gene2vec}, multi-hot encoded Reactome (\url{https://reactome.org}), and STRING embeddings \cite{szklarczyk2019string}.

Foundational models are the relatively new line of work based on Transformer neural network architecture \cite{vaswani2017attention}. Transformer-based architectures are used to learn gene embeddings from abundant observational gene expression data using various strategies for tokenising transcript counts. These models (frequently referred to as \enquote{foundation models}) include Geneformer \citep{theodoris2023transfer}, scGPT \citep{cui2023scgpt}, scFoundation \citep{hao2023large}, and GenePT \citep{chen2023genept}. Unlike LPM, Transformer-based methods are trained on observational data and hence lack the data to make transportable causal inferences, and furthermore produce only gene embeddings that cannot directly be used to understand pharmacological perturbations. Note also that it is not established whether the tokenisation required by Transformer-based models leads to performance improvements over more straightforward representations of tabular experimental transcriptomics data. The contrast between foundational models and \themethod{} has been in more detailed covered in the introduction of this paper.

Worth-noting are a sequence of methods that have a causal perspective to perturbation modeling.
CINEMA-OT \cite{dong2023causal} and Cell-OT \cite{bunne2023learning} use optimal transport to infer treatment effects and predict responses of single cells to genetic or drug-based perturbations, with a particular focus on pairing (matching) intrinsically unpaired distributions of perturbed and non-perturbed cells.
CINEMA-OT attempts to explicitly separate confounding sources of variation from perturbation effects.
The advantage of these methods is their ability to predict post-perturbation outcome for single cells (individual treatment-effect analysis).
sVAE+ \cite{lopez2020enhancing} is a variant of the sVAE with a Bayesian aproach for learning sparse mechanism shifts.
sVAE+ models perturbations in single-cell space as unknown but sparse interventions on latent space, using concepts of sparsity to enhance its utility.
sVAE+ offers interpretability features and covers novel aspects such as latent units recovery and causal target identification.
These methods are not designed for learning from multi-context data.

Understanding the generalizability of perturbation models necessitates examining their capabilities in handling in-vocabulary (IV) and out-of-vocabulary (OOV) predictions.
\themethod{}, as presented here, operates entirely within an IV framework, meaning they can predict only combinations of symbols encountered during training.
While this may seem limiting, it's important to note: \emph{(i)} The number of perturbation studies is steadily increasing, suggesting that the symbolic space will soon be substantially covered; \emph{(ii)} There's potential to replace randomly initialized embeddings with predefined ones, similar to approaches like GEARS, to facilitate OOV predictions for perturbations, contexts, or readouts.
Variational Autoencoder (VAE)-based approaches are also predominantly IV. However, unlike LPM, and with the exception of models like ChemCPA, they often struggle with heterogeneity in multi-context settings.
GEARS can predict effects of OOV perturbations, limited to symbols present in the Gene Ontology, but it cannot perform OOV context predictions.
Foundational models like scGPT are capable of IV perturbation prediction and OOV context prediction.
In summary, while LPM is currently restricted to IV predictions, the expanding dataset landscape and potential integration of predefined embeddings offer ways to enhance their OOV capabilities.

\section{Other results}
\label{sec:other-results}

\paragraph{Extended results from the benchmarks presented in the main part of the paper.}

\begin{figure*}[!p]
\centering
\includegraphics[width=\textwidth]{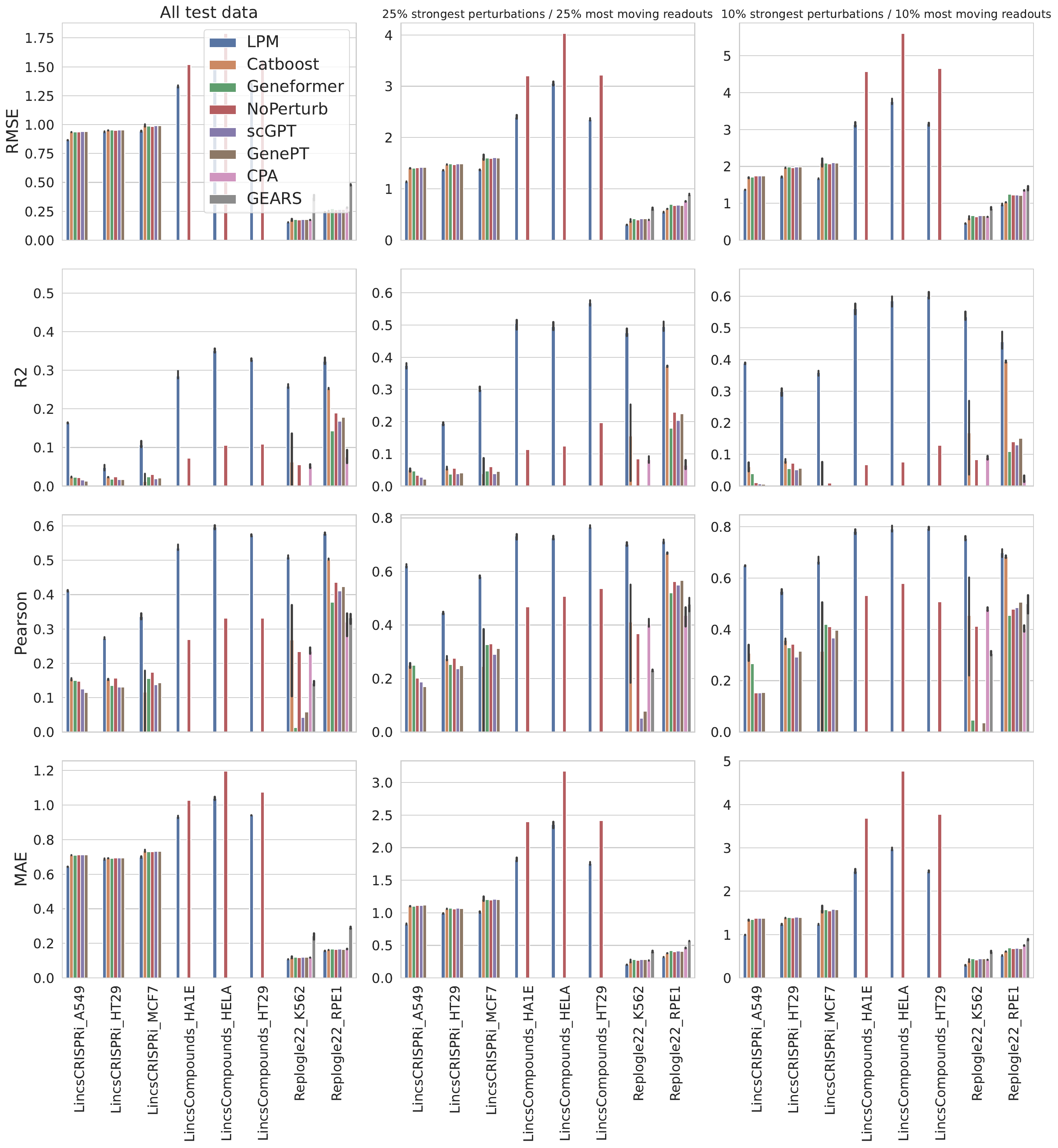}
\caption{\textbf{Detailed performance comparison in predicting unperformed experiments.} {\themethod} consistently performs better across considered datasets, metrics, and with respect to different portions of the test set in predicting outcomes of unobserved perturbations. Given that the transfer learning setting studied here consists of only one dataset, we expect performance could potentially improve as we include more datasets in the training process.}
\label{fig:quantitative-analysis}
\end{figure*}

The detailed predictive performance benchmark is given in \cref{fig:quantitative-analysis}.
For the task of gene regulatory network inference comparison, we present the full performance statistics for the gene regulatory network inference task outlined in \cref{sec:genenetworkinference} in \cref{tab:ranking_k562_50}. %

\begin{table}[!h]
    \centering
    \begin{tabular}{lrrrrr}
    \toprule
    \multicolumn{6}{c}{K562 cells} \\
    Model                    & Mean & Wasserstein & FOR  & Wasserstein   & FOR           \\
                             & Rank     & Rank        & Rank & Distance      &               \\
    \midrule
    Mean Difference (top 5k) & 3.0 & 4 & 2 & 0.371 ± 0.005 & 0.158 ± 0.004 \\
    \textbf{\themethod{}+Guanlab}  & 4.0 & 7 & 1 & 0.318 ± 0.005 & 0.150 ± 0.005 \\
    Guanlab (top 1k) & 4.0 & 2 & 6 & 0.430 ± 0.010 & 0.181 ± 0.001 \\
    Mean Difference (top 1k) & 4.0 & 1 & 7 & 0.523 ± 0.007 & 0.182 ± 0.003 \\
    Guanlab (top 5k) & 4.5 & 6 & 3 & 0.327 ± 0.005 & 0.161 ± 0.002 \\
    Moebius & 5.0 & 5 & 5 & 0.341 ± 0.002 & 0.176 ± 0.003 \\
    Betterboost & 5.5 & 3 & 8 & 0.420 ± 0.006 & 0.186 ± 0.001 \\
    NoPerturb+Guanlab & 6.0 & 8 & 4 & 0.266 ± 0.004 & 0.161 ± 0.004 \\
    \bottomrule \\
    \toprule
    \multicolumn{6}{c}{RPE1 cells} \\
    Model                    & Mean & Wasserstein & FOR  & Wasserstein   & FOR           \\
                             & Rank     & Rank        & Rank & Distance      &               \\
    \midrule
    Moebius & 3.5 & 4 & 3 & 0.293 ± 0.003 & 0.115 ± 0.004 \\
    \textbf{\themethod{}+Guanlab} & 4.0 & 7 & 1 & 0.274 ± 0.006 & 0.087 ± 0.008 \\
    Guanlab (top 1k) & 4.0 & 2 & 6 & 0.496 ± 0.008 & 0.126 ± 0.003 \\
    Mean Difference (top 5k) & 4.5 & 5 & 4 & 0.289 ± 0.004 & 0.116 ± 0.005 \\
    Mean Difference (top 1k) & 4.5 & 1 & 8 & 0.556 ± 0.009 & 0.142 ± 0.003 \\
    NoPerturb+Guanlab & 5.0 & 8 & 2 & 0.260 ± 0.009 & 0.091 ± 0.008 \\
    Betterboost & 5.0 & 3 & 7 & 0.474 ± 0.011 & 0.132 ± 0.004 \\
    Guanlab (top 5k) & 5.5 & 6 & 5 & 0.276 ± 0.004 & 0.119 ± 0.004 \\
    \bottomrule
    \end{tabular}
    \caption{Complete gene network inference performances on the K562 (top) and RPE1 (bottom) cell lines with  50\% of interventional data. }
    \label{tab:ranking_k562_50}
\end{table}

\paragraph{Additional results on post-perturbation prediction.}

\begin{figure*}[!h]
\centering
\includegraphics[width=0.5\textwidth]{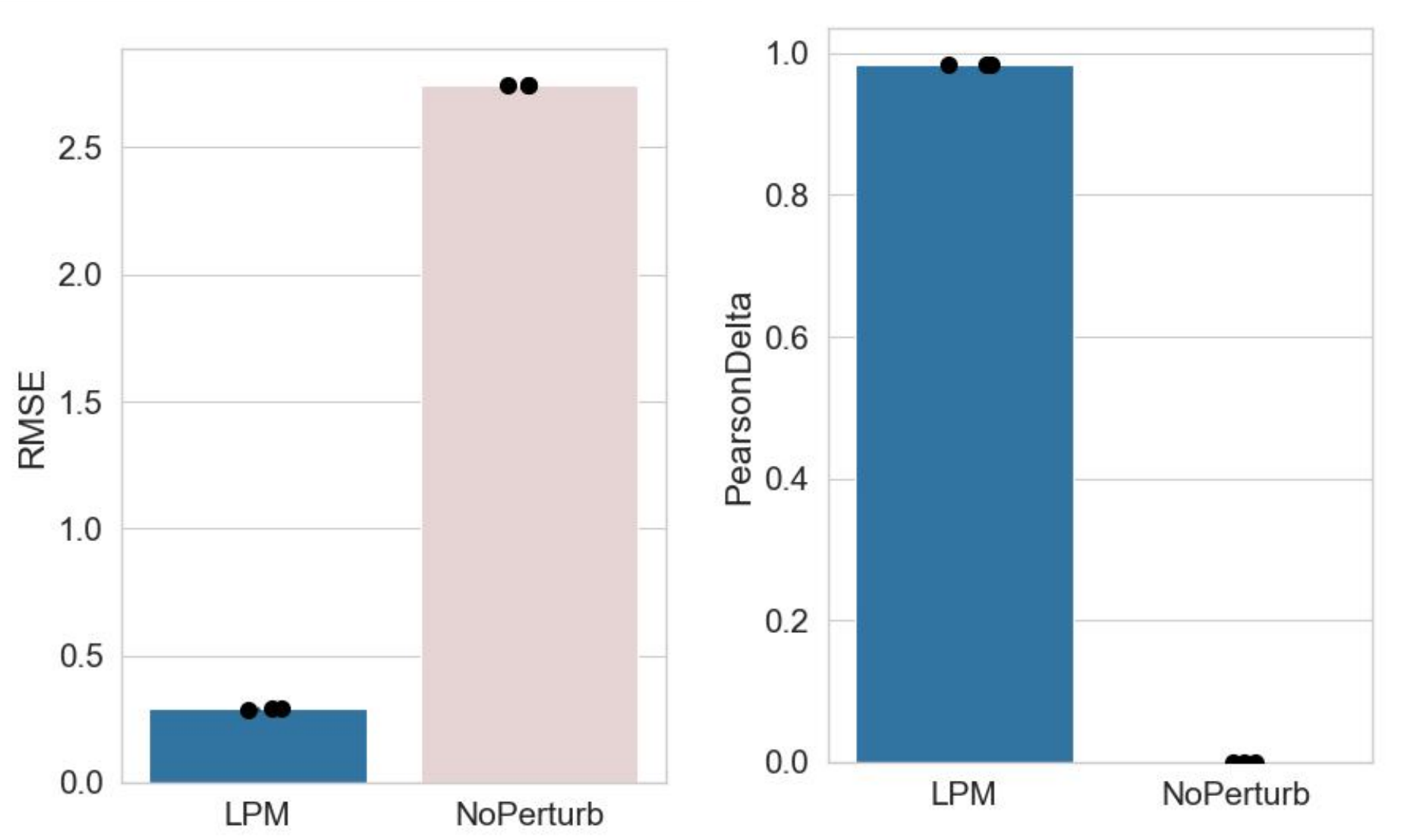}
\caption{\textbf{Predicting unperformed experiments on perturbation screens that measure cell viability \cite{horlbeck2018mapping}.} {\themethod} shows considerable performance improvements in comparison to the simple NoPerturb baseline.}
\label{fig:viability-experiments}
\end{figure*}

We conducted an additional experiment on the effects of CRISPRi genetic perturbations on cell viability, based on the study by \cite{horlbeck2018mapping}, which explored how genetic interactions impact viability.
This dataset includes three cell viability-related readouts derived from pairwise perturbations across two cell types, demonstrating the adaptability of our method to data beyond transcriptomics and in low-dimensional settings.
Our experimental setup followed the approach described above: each single perturbation involved in a pairwise combination was included during training, while the specific combinations themselves were withheld.
Note that because we used only the average of two replicates, GEARS was not applicable in this case, as it currently supports only single-cell data.
Similarly, scGPT could not be applied due to the non-transcriptomic nature of this dataset.
Results are shown in \cref{fig:viability-experiments}.

\end{document}